\documentclass[10pt,twocolumn,letterpaper]{article}

\usepackage{3dv}
\usepackage{times}
\usepackage{epsfig}
\usepackage{graphicx}
\usepackage{amsmath}
\usepackage{amssymb}
\usepackage{mathtools}
\usepackage{enumitem}
\usepackage{rotating}
\usepackage{multirow}
\usepackage{colortbl}
\usepackage[dvipsnames]{xcolor}
\usepackage{adjustbox}
\usepackage{array}
\newcolumntype{R}[2]{
    >{\adjustbox{angle=#1,lap=1.3\width-(#2)}\bgroup}%
    l
    <{\egroup}
}
\newif\ifarxiv
\arxivfalse

\usepackage{times}
\usepackage{epsfig}
\usepackage{graphicx}
\usepackage{amsmath}
\usepackage{amssymb}
\usepackage{mathtools}
\usepackage{enumitem}
\usepackage{rotating}
\usepackage{adjustbox}
\usepackage{array}
\usepackage[dvipsnames]{xcolor}
\usepackage{colortbl}
\usepackage{color}
\usepackage{booktabs}
\usepackage{multirow}
\usepackage{makecell}

\definecolor{HL}{rgb}{0.95,1,0.95}
\definecolor{bisque}{rgb}{1.0, 0.89, 0.77}
\definecolor{blanchedalmond}{rgb}{1.0, 0.92, 0.8}
\definecolor{palepink}{rgb}{0.98, 0.85, 0.87}
\definecolor{piggypink}{rgb}{0.99, 0.87, 0.9}
\definecolor{lightred}{rgb}{1, 0.93, 0.93}
\definecolor{LightGrey}{rgb}{0.95,0.95,0.95}

\newcommand{\class}[1]{`\emph{#1}'}
\newcommand{\seen}{\mathcal{S}}
\newcommand{\unseen}{\mathcal{U}}
\newcommand{\semantic}{\mathcal{T}}
\newcommand{\pointcloud}{\mathcal{P}}

\newcommand{\classes}{\mathcal{C}}
\newcommand{\labels}{\mathcal{Y}}

\newcommand{\test}{\text{test}}

\newcommand{\abstracttext}{
While there has been a number of studies on Zero-Shot Learning (ZSL) for 2D images, its application to 3D data is still recent and scarce, with just a few methods limited to classification. We present the first generative approach for both ZSL and Generalized ZSL (GZSL) on 3D data, that can handle both classification and, for the first time, semantic segmentation. We show that it reaches or outperforms the state of the art on ModelNet40 classification for both inductive ZSL and inductive GZSL. For semantic segmentation, we created three benchmarks for evaluating this new ZSL task, using S3DIS, ScanNet and SemanticKITTI. Our experiments show that our method outperforms strong baselines, which we additionally propose for this task.}
\arxivtrue

\usepackage[pagebackref=true,breaklinks=true,colorlinks,bookmarks=false]{hyperref}

\ifarxiv
  \threedvfinalcopy
\fi

\def\threedvPaperID{****}

\ifthreedvfinal\fi
\begin{document}

\title{Generative Zero-Shot Learning for Semantic Segmentation of 3D Point Clouds}

\author{Bj\"orn Michele\textsuperscript{1}
\and
Alexandre Boulch\textsuperscript{1}
\and
Gilles Puy\textsuperscript{1}
\and 
Maxime Bucher\textsuperscript{1}
\and 
Renaud Marlet\textsuperscript{1, 2}
\and
\\
\large
\hspace{-3mm}\textsuperscript{1}Valeo.ai, Paris, France  \hspace{1mm}
\textsuperscript{2}LIGM, Ecole des Ponts, Univ Gustave Eiffel, CNRS, Marne-la-Vall\'ee, France
}

\maketitle

\begin{abstract}
\abstracttext
\end{abstract}

\section{Introduction}

3D perception is a critical part of many applications.
We consider here two perception tasks on 3D point clouds: classification and, more importantly, semantic segmentation.
The state of the art for these tasks is currently achieved by deep nets trained under full supervision.
Yet, while 3D sensors have become more affordable, labeling 3D data has remained costly and time consuming. Semantic segmentation datasets for point clouds therefore contain a limited number of object and scene classes, with little intra-class variation, thus only covering partial real world situations.
An option to address these limitations is to try to make predictions at inference time for objects unseen at training time, based on auxiliary information regarding non-annotated classes.

Zero-Shot Learning (ZSL) only predicts classes unseen at training time; Generalized ZSL (GZSL) predicts both seen and unseen classes.
More precisely, while transductive (G)ZSL allows unlabeled objects of unknown classes to be part of training data, inductive (G)ZSL forbids it, making objects of unknown classes totally new to the model.

\begin{figure}
    \centering
    \begin{tabular}{@{}c@{\hskip 0.05cm}c@{\hskip 0.15cm}c@{\hskip 0.05cm}c@{}}
        \includegraphics[trim=0 25 0 25,clip,width=0.24\linewidth]{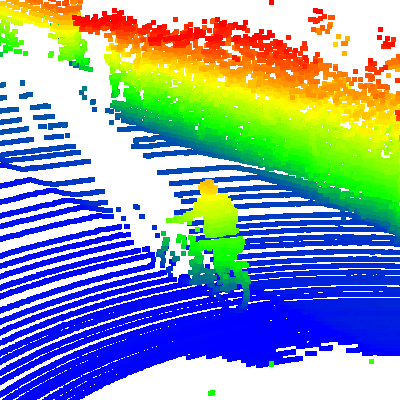}&
        \includegraphics[trim=0 25 0 25,clip,width=0.24\linewidth]{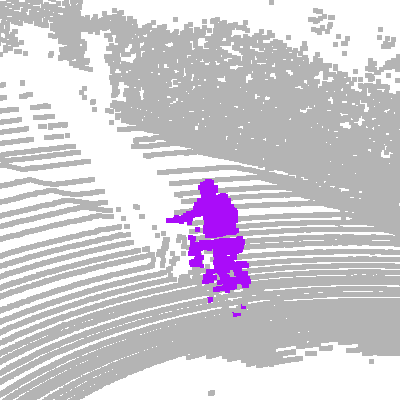}&
        \includegraphics[trim=0 25 0 25,clip,width=0.24\linewidth]{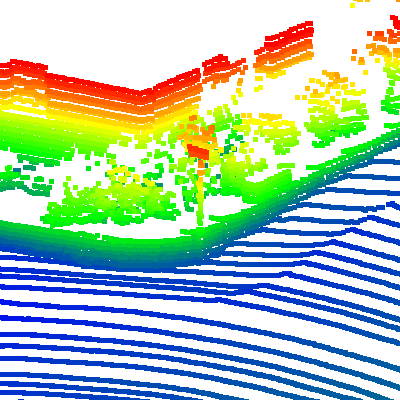}&
        \includegraphics[trim=0 25 0 25,clip,width=0.24\linewidth]{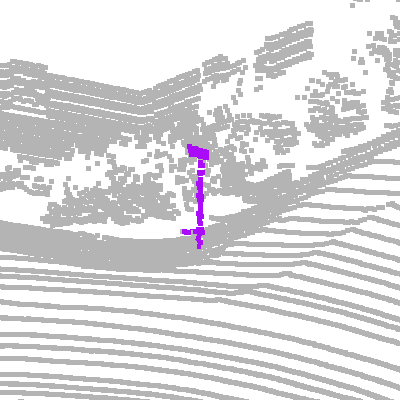}\\
        \includegraphics[trim=0 25 0 25,clip,width=0.24\linewidth]{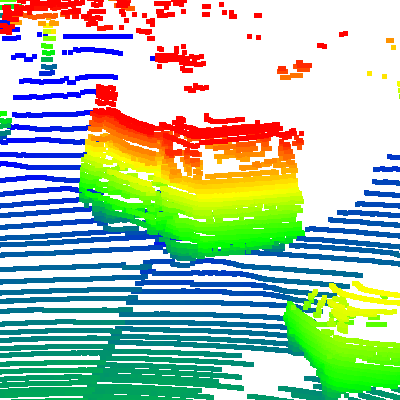}&
        \includegraphics[trim=0 25 0 25,clip,width=0.24\linewidth]{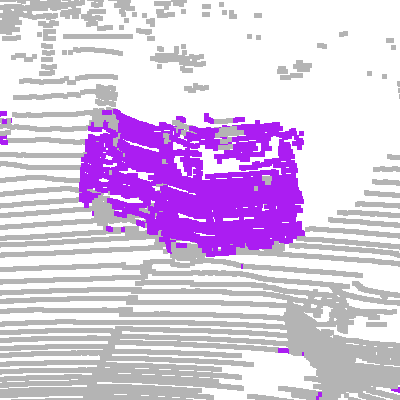}&
        \includegraphics[trim=0 25 0 25,clip,width=0.24\linewidth]{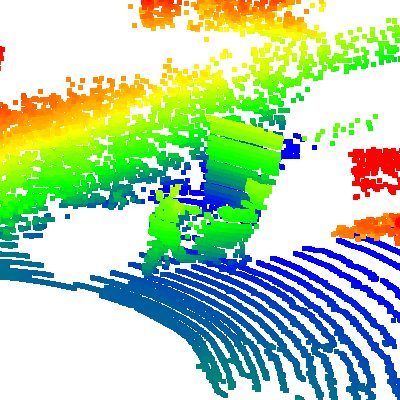}&
        \includegraphics[trim=0 25 0 25,clip,width=0.24\linewidth]{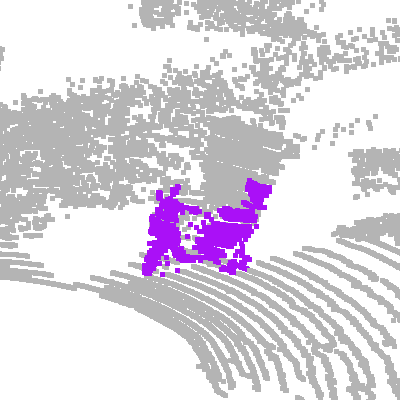}
    \end{tabular}
    \caption{Zero-shot point cloud segmentation on Semantic\-KITTI. Point cloud with color gradient according to height (left image) and ZSL segment focusing on an unseen class (right) for classes bicyclist, traffic sign (top row), truck and motorbike (bottom row)}
    \label{fig:teaser}
    \vspace*{-1mm}
\end{figure}

Much progress has been made on ZSL for image classification~\cite{rezaei2020zero,xian2018zero,wang2019survey} and, recently, semantic segmentation \cite{bucher2019zero, xian2019semantic, hu2020uncertainty, li2020consistent, cheng2021sign}.
But ZSL for point clouds has only been investigated for classification and by few studies \cite{cheraghian2019mitigating,cheraghian2019zero,cheraghian2020transductive}.
We present here the first (to our knowledge) 3D GZSL approach for semantic segmentation.

Classification makes sense for individual objects that are more or less isolated or centered. But except when making a digital 3D copy of an object (an easy labeling time), scans often observe a complex scene rather than a single object. 3D classification remains however relevant in the object detection task, when applied to a region proposal. We believe ZSL semantic segmentation is a more realistic scenario, applying to complex outdoor or indoor scenes as scanned by lidars or range cameras. It can be useful in particular as a 3D pre-annotation tool, e.g., for autonomous driving where country-specific objects (vehicles, roadsides, work barriers, possible road obstacles\ldots) have to be widely collected and labeled. More precisely, the most relevant task is not ZSL but GZSL semantic segmentation, as it makes more sense not to forget about known classes when pre-annotating. Zero-shot segmentation can also be useful in the few-shot scenario as a way to mine large point cloud datasets to retrieve some examples to be manually annotated.

Our contributions are as follows. (1)~We propose a generative framework handling both ZSL and GZSL for 3D point clouds, both for semantic segmentation  and classification.  (2)~We make available 3 benchmarks for 3D GZSL semantic segmentation based, indoors, on S3DIS \cite{armeni20163d} and ScanNet \cite{dai2017scannet}, and outdoors, on SemanticKITTI~\cite{behley2019iccv} (cf.\ Fig.\,\ref{fig:teaser}). (3)~We define 2 baselines for 3D GZSL segmentation, which our method outperforms on the benchmarks.
\section{Related work}
\label{sec:related}

\subsection{Zero-shot learning for images}
\label{sec:related:zsl}

ZSL can be viewed as a special case of transfer learning, where knowledge from a source domain (seen classes) and source task (classification or segmentation) is transferred to a target domain (unseen classes) with a target task (different label space) \cite{rezaei2020zero,xian2018zero,wang2019survey}. We review some existing methods.

\textbf{Attribute classifier.~}
A first class of methods aims to recognize new objects based on attributes
\cite{lampert2009learning, farhadi2009describing, liu2011recognizing, jayaraman2014zero, al2016recovering}.
Here, no attribute list is available.
But methods like ConSE~\cite{norouzi2013zero} can also use word embeddings as attributes and can thus be adapted for ZSL on 3D point clouds~\cite{cheraghian2019zero}.

\textbf{Projection methods.~}
This class of methods creates a mapping between an object representation and auxiliary data (class prototypes) such as word embeddings, e.g., W2V~\cite{mikolov2013distributed} or GloVe~\cite{pennington2014glove}.
Class are then assigned in the prototype space \cite{frome2013devise,xian2016latent,socher2013zero}.
However, these methods often face the hubness problem~\cite{radovanovic2010hubs}, where a (hub) class prototype is the nearest neighbor of a large number of other prototypes.
To tackle this problem, an alternative is to do the comparison in the object representation space~\cite{shigeto2015ridge, zhang2017learning}.

\textbf{Generative models.~}
ZSL can be seen as a missing data problem: no examples of unseen classes are available at training time.
\emph{Generative methods} create artificially this missing data to train a classifier under supervision~\cite{Bucher_2017_ICCV,xian2018feature}.
As in~\cite{frome2013devise}, a CNN extracts visual features of seen classes, which are used to train the generative module, conditioned on the corresponding class prototype.
Generative models are known to reduce the bias towards seen classes in GZSL and often to be superior to projection methods~\cite{schonfeld2019generalized}.
A great variety of generative modules may be used to create artificial features. Adversarial auto-encoders~\cite{makhzani2015adversarial}, conditional generative adversarial models~\cite{odena2017conditional}, denoising auto-encoders~\cite{bengio2013generalized} or GMMNs~\cite{li2015generative} are used in~\cite{Bucher_2017_ICCV}.
Wasserstein GANs~\cite{gulrajani2017improved} have been used in~\cite{xian2018feature}, and VAEs in \cite{kumar2018generalized,xian2019f}.

While f-CLSWGAN~\cite{xian2018feature} focuses on the GAN aspect to make generated features somehow look more realistic and only uses the classification loss as a regularizer, our feature generation is only driven by classification (as~\cite{bucher2019zero}), which is the task target anyway. We thus save the tuning of 2 extra hyperparameters and we do not face the difficulty to train a GAN, as recognized in~\cite{xian2018feature}. Besides, conditioning a discriminator on word embeddings is probably harder than on the simple attributes used in~\cite{xian2018feature} itself. Moreover, training a discriminator to make generated features look like real features may be harmful with 3D datasets, that offer less training data than image datasets, thus less samples to learn real-looking features. The fact is we significantly outperform the adpatation of f-CLSWGAN 3D (cf.\ Table~\ref{tab:zsl_gzsl_overview}).

\textbf{Semantic segmentation.~}
While ZSL for image classification has been studied extensively (see above), semantic segmentation has only recently been tackled.
\cite{zhao2017open,kato2019zero} focus on the discovery of objects of interest in a scene, either using a hierarchical open vocabulary approach~\cite{zhao2017open} or splitting semantic segmentation into a foreground/background segmentation step and a classification step~\cite{kato2019zero}.

Other methods address the problem in a GZSL setting. \cite{xian2019semantic,hu2020uncertainty} project the object representation onto the class prototype space using semantic projection layers.
On the contrary, \cite{bucher2019zero, gu2020context,li2020consistent,cheng2021sign} project class prototype representations onto the object representation space and generate pixel-wise features of seen and unseen classes that are used to train a classifier. Our generative method belongs to this second group of approaches, adapting them to the special case of 3D point clouds.

\subsection{Point cloud analysis with deep learning}
\label{sec:point_cloud_processing}

Simple ways to adapt 2D methods to 3D data include conversion to range images \cite{gupta2014learning, long2015fully}, image generation from virtual viewpoints \cite{su2015multi, boulch2017snapnet, lawin2017deep}, projection on 2D planes~\cite{tatarchenko2018tangent} and using voxel grids~\cite{maturana2015voxnet,roynard2018classification,qi2016volumetric,wu20153d, riegler2017octnet, graham20183d}.
Graph Neural Networks (GNNs) have been used to limit the loss of information due to data projection. They operate on graphs based on 3D neighborhoods \cite{scarselli2008graph,bronstein2017geometric}, possibly pre-segmented~\cite{landrieu2018large},
using message passing~\cite{gilmer17,li2015gated} or defining convolution in the spectral domain \cite{bruna14,defferrard2016convolutional,kipf17}.
Deep learning on raw point clouds has now become commonplace. The points can be processed all together, like in PointNet~\cite{qi2017pointnet}, or using a hierarchical structure~\cite{qi2017pointnet++,li2018so,hua2018pointwise,thomas2019kpconv,boulch2020convpoint,xu2018spidercnn,wang2018deep,liu2019relation,wu2019pointconv}.

\subsection{Zero-shot learning for 3D point clouds}

To our knowledge, only 4 publications study ZSL for point clouds \cite{cheraghian2019zero,cheraghian2019mitigating, cheraghian2020transductive, cheraghian2021zero}, and they only address the classification task.  
The pioneering work \cite{cheraghian2019zero} adapts ConSE \cite{norouzi2013zero} to 3D, using PointNet~\cite{qi2017pointnet} to create an object representation, and GloVe \cite{pennington2014glove} or W2V \cite{mikolov2013distributed} as auxiliary information.
\cite{cheraghian2019mitigating} reduces the hubness problem of \cite{cheraghian2019zero} using a loss function composed of a regression term~\cite{zhang2017learning} and a skewness term~\cite{radovanovic2010hubs, shigeto2015ridge}, and extends to GZSL.
The transductive case is discussed in~\cite{cheraghian2020transductive} which extends~\cite{cheraghian2019mitigating} using a triplet loss. Finally, the hubness problem is addressed in~\cite{cheraghian2021zero} along with the proposition of a unified approach for~\cite{cheraghian2019zero, cheraghian2019mitigating, cheraghian2020transductive}.

None of these approaches is generative. Yet, \cite{cheraghian2020transductive} transposes generative 2D methods \cite{xian2018feature,schonfeld2019generalized} to 3D for comparison. Poor results lead \cite{cheraghian2020transductive} to hypothesize they do not generalize well to 3D because their performance in 2D is mostly due to the high quality of pre-trained models (on millions of labeled images featuring thousands of classes), which do not exist for 3D data.
In this work, we show that even with small datasets and a few classes, generative methods outperform the state of the art for 3D point clouds ZSL and GZSL, and also generalize to 3D semantic segmentation.
Recently, semantic segmentation is discussed in the technical report \cite{liu2021segmenting}, concurrent to our work. 
The setting used in~\cite{liu2021segmenting} can be referred as zero-label learning~\cite{Xian_2019_CVPR}, i.e., the unseen classes are present at training time.
On the contrary, we follow a ZSL procedure, removing from the training set all point clouds containing unseen classes. Furthermore, we are also addressing in our method the bias problem, which semantic segmentation faces due to the appearance of seen and unseen classes in the same scene.

\section{Method}
\label{sec:method}

Point cloud semantic segmentation can largely be seen as the classification of individual 3D points, although specific developments are required (see below). In this section, we introduce a general generative ZSL framework that applies both to classification and semantic segmentation.

\subsection{Problem formulation}
\label{sec:problem}

Let $\classes$ be a set of object classes, $\pointcloud = (P_i)_{i\in I}$ be a set of objects, and $\labels= (y_i)_{i\in I}$ be the set of corresponding class labels. For classification, an \emph{object} $P_i$ to label is a point cloud; for semantic segmentation, it is a 3D point.
    
The object classes $\classes$  are partitioned into seen classes $\seen$ and unseen classes $\unseen$.
The objects of seen classes are $\pointcloud^\seen = (P_i)_{i\in I^\seen}$ with $I^\seen = \{i\in I \mid y_i \in \seen\}$, and likewise for unseen classes $\unseen$. At training time, only objects $(P_i)_{i\in I^\seen}$ and corresponding class labels $(y_i)_{i\in I^\seen}$ are available.

\textbf{Class prototypes.}
Learning from $\pointcloud^\seen$ and generalizing to $\pointcloud^\unseen$ without seeing any example from a class in $\unseen$ is impossible without extra knowledge.
We have to rely on \emph{auxiliary information}: the so-called \emph{class prototypes}, which are not exemplars (as no ``shot'' is allowed) but $D$-dimensional embedding vectors. It is denoted by $\semantic = \{t_c \in \mathbb{R}^D \mid c \in \classes\}$, where each class has a single  prototype.
We distinguish $\semantic^\seen$ and $\semantic^\unseen$, the subsets of $\semantic$ for seen and unseen classes.
    
\textbf{Object representations.}
Our objective is to embed the objects in $\pointcloud$ and the class prototypes in $\semantic$ into a common object representation space $\mathcal{X}$ where objects and prototypes of the same class have similar embeddings. The embedding function for objects is denoted by $\phi(\cdot)$ and is typically implemented using a deep neural network. Our embedding function for class prototypes is a generator denoted by $G(\cdot)$.

\textbf{Training set.}
We consider the difficult case of \emph{inductive} ZSL: no data on unseen classes is available at training time; only their class prototypes is available, at test time.
The training set thus consists of the triplets $(P_i, y_i, t_{y_i})_{i\in I^\seen}$ where $(t_{y_i})_{i\in I^\unseen}$ are the class prototypes of unseen classes.

\textbf{Test set.}
We test on objects $\pointcloud_\test \,{=}\, (P_i)_{i\in I_\test}$ labeled in $\labels_\test \,{=}\, (y_i)_{i\in I_\test}$, where $I_\test$ indexes test samples.
When $\labels_\test$ contains only classes in $\unseen$, it is the \emph{vanilla ZSL} test setting;
when $\labels_\test$ contains classes both in $\seen$ and $\unseen$, it is the \textit{generalized ZSL} test setting.
Semantic segmentation mainly makes sense in complex scenes with several co-located objects of different classes. As in practice, seen and unseen classes will often simultaneously appear in a scene, we consider semantic segmentation only in the GZSL setting.

\subsection{Our approach}
\label{sec:approach}

\begin{figure}
    \centering
    \includegraphics[width=\linewidth]{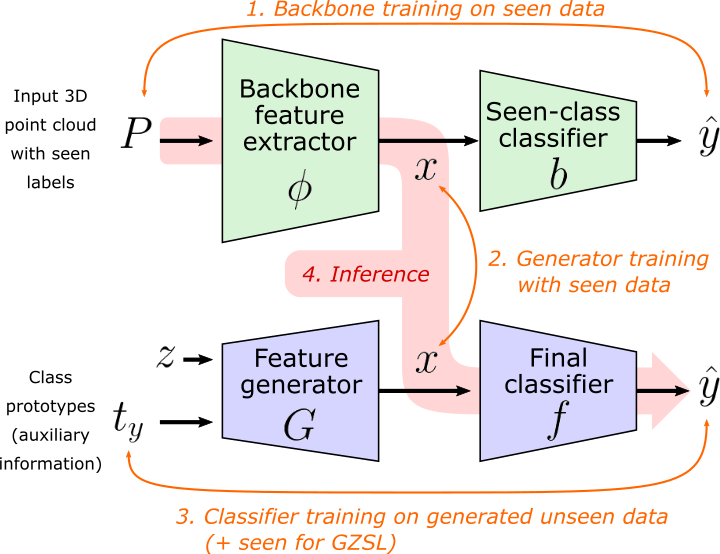}
    \caption{Four-step training and inference procedure: (1)~backbone training on the seen classes, (2)~generator training, (3)~classifier training with artificial unseen features for ZSL (unseen and seen for GZSL), (4)~inference through backbone and final classifier.}
    \label{fig:approach}
\end{figure}

Our approach relies on 3 main modules, trained sequentially (see Fig.\,\ref{fig:approach}): a backbone feature extractor $\phi(\cdot)$ processing 3D point clouds, a generative module $G(\cdot)$ taught to generate features $x$ conditioned on a class prototype $t$, and a classifier $f(\cdot)$ predicting a class $\hat{y}$ given a feature~$x$.

\textbf{Backbone feature extractor.}
\label{sec:approach:backbone}
Each object $P_i$ in $\pointcloud$ is represented by a set of 3D points $\bar{P_i}$. For classification, $\bar{P_i}$ is the point cloud $P_i$ itself; for semantic segmentation, $\bar{P_i}$ is the point cloud containing the point $P_i$ to label.
The object embedding function $\phi(\cdot)$ is a backbone, such as PointNet \cite{qi2017pointnet}. It extracts one representation $x = \phi(P_i)$ for classification, or several (one for each point of $\bar{P_i}$) for segmentation.

We first train the feature backbone $\phi(\cdot)$ under full supervision of the seen classes, combining it with a linear classifier $b(\cdot)$. For each example $(P_i,y_i)_{i\in I^\seen}$ of a seen class, we compare $b(\phi(P_i))$ to $y_i$ via a cross-entropy loss to train both $b(\cdot)$ and $\phi(\cdot)$. Only $\phi(\cdot)$ is used afterwards; $b(\cdot)$ is ignored.

\textbf{Feature generator.}
\label{sec:approach:generator}
The role of generator $G(\cdot)$ is to create a training set of fake but realistic unseen object representations, to train a final classifier for unseen classes. These generated features should be similar to $\phi(P)$ for $P \in \pointcloud^\unseen$, i.e., to features obtained if we had had access to unseen data. For this, we use a generative approach like in \cite{Bucher_2017_ICCV, bucher2019zero}. Unlike $\phi$, that is deterministic, features of class prototypes are generated as $x\,{=}\, G(\mathbf{z}_j, t_{s})$, where $\mathbf{z}_j$ is a random vector and $t_s \,{\in}\, \semantic^\seen\!$ is the prototype of a seen class $s \,{\in}\, \seen$. $G(\cdot)$ is trained to generate features $x_i \,{=}\, G(\mathbf{z}_j, t_{y_i})$ similar to backbone features $\phi(P_i)$, from examples $(P_i,y_i)_{i\in I^\seen}$ of seen classes.

\textbf{Final classifier.}
\label{sec:approach:classifier}
The final feature classifier $f(\cdot)$ is trained with a cross-entropy loss. For ZSL, it supervised by generated features of unseen classes only, i.e., using a training set $D^\unseen = \{(G(\mathbf{z}_j, t_y), y) \mid y \in \unseen, 1 \leq j \leq \vert D^\unseen\vert\}$. For GZSL, $f(\cdot)$ is trained on $D^{\unseen \cup \seen}$,
containing features generated for unseen classes and features of the seen training data (backbone outputs). Classification is successful if the distribution of generated features in $D^\unseen$ is similar to that of $\phi(P)$ for $P$ in $\pointcloud^\unseen$ (or $P$ in $\pointcloud^\classes$ for GZSL) and if the object representation space can be linearly separated for each class.

\textbf{Inference.}
For inference, given a point cloud $P$, features are first extracted using the trained backbone $\phi$, then classified using the final classifier $f$ (see Fig.\,\ref{fig:approach}). In other words, inference consists in computing $\hat{y} = f(\phi(P))$.
Therefore, the inference time and space complexity basically is that of the backbone. The generator is only used at training time.

\textbf{Reducing bias toward seen classes.}
In GZSL, both unseen and seen classes appear in the test set.
As described in \cite{chao2016empirical}, a bias toward seen classes can then
be observed.
The reason is twofold.
First, the feature extractor may focus only on features useful to discriminate seen classes,
inducing a loss of information required to deal with unseen classes.
Second, as the generator is trained only on seen classes, it generates feature of better quality for seen classes than unseen ones. 
This bias is addressed in two ways:

\emph{Class-dependent weighting.~}
When we train the classifier, the loss for unseen classes is weighted with a factor $\beta>1$.
The assumption is that, as the generator is only trained on seen classes, it generates lower quality features for unseen classes, which are thus more difficult to classify.
Therefore, we give more importance to the unseen classes at training time, forcing the classifier to focus on them.

\emph{Calibrated stacking}~\cite{chao2016empirical}.~
The bias for seen classes can be reduced at test time as a post-processing by subtracting a small value $\epsilon$ from the seen-class score (after softmax).

The weight factor $\beta$ and the offset $\epsilon$ are hyperparameters chosen using a validation set created from the train set.

\section{Experiments}
\label{sec:experiments}

\subsection{Datasets and Metrics}

\textbf{Classification.}
ModelNet40 \cite{wu20153d} is used in \cite{cheraghian2019zero, cheraghian2019mitigating, cheraghian2020transductive, cheraghian2021zero} as a classification benchmark.  It consists of  40 object classes of CAD objects. For the ZSL setting, it is split into 30 seen and 10 unseen classes. The 10 unseen classes are the ones of ModelNet10, which is a subset of ModelNet40.

\textbf{Semantic segmentation.}
As it is the first time ZSL semantic segmentation is tackled for 3D data, there is no reference benchmark. We created 3, based on 3 common 3D semantic segmentation datasets.
S3DIS \cite{armeni20163d} includes point clouds of 271 scanned rooms, with points labeled among 13 classes.
ScanNet \cite{dai2017scannet} contains 1513 indoor scans with annotations for 20 classes.
SemanticKITTI \cite{behley2019iccv,Geiger2012CVPR}
provides point clouds acquired by a lidar on a car driving in the streets. Grouping moving and non-moving objects with the same semantics results in 19 different classes.
We keep the same 10 sequences for training but, as test sequences are unavailable, we use the validation sequence for ZSL testing.

\textbf{ZSL splits.}
To assess ZSL, we need seen and unseen classes.
To ease the adoption of our benchmark, we create a single but rich ZSL split per dataset, with a variety of difficulties, while allowing to leverage on textual semantic proximity. 
To make the datasets appropriate for the inductive setting, we discard, in the seen-class training set, any point cloud containing an instance of an unseen class.
It is a hard but necessary constraint, although it reduces the size of the original datasets. The fact is keeping all point clouds in their entirety and learning only from labeled seen-class points cannot qualify as an inductive setting because the backbone can then leverage on the contextual presence of unlabeled unseen-class points, even if these unseen-class points do not back-propagate class information.

The small number of classes and their distribution reduce options as there must be enough samples of seen classes to train on.
Yet to make segmentation challenging we consider 4 unseen classes in each dataset:
\emph{beam}, \emph{column}, \emph{window}, \emph{sofa} for S3DIS; \emph{desk}, \emph{bookshelf}, \emph{sofa},  \emph{toilet} for ScanNet; \emph{motorcycle}, \emph{truck}, \emph{bicyclist}, \emph{traffic-sign} for SemanticKITI.
For indoor scenes (S3DIS, ScanNet), expected semantically close categories are \emph{sofa} (unseen) and \emph{chair} (seen), and \emph{desk} (unseen) and \emph{table} (seen).
For outdoors (SemanticKITTI), \emph{poles} are chosen as seen while \emph{traffic signs}, most of which are attached to a pole, are not. To allow inductive ZSL, traffic-sign poles are thus not seen at training time. Yet the split allows to evaluate the correlation. Moreover, \emph{bicycles} (usually parked) are seen while \emph{bicyclists} (including the bikes) are not; it allows to evaluate the ability to add an unseen rider onto a seen class, and conversely for \emph{motorcycle} and \emph{motorcyclist}.
Last, \emph{trucks} are unseen, with no direct correlation other than being vehicles like \emph{cars}.

\textbf{Metrics.}
We evaluate the methods with commonly used metrics: global accuracy (Acc) and accuracy per class for classification; average intersection-over-union (mIoU) for semantic segmentation.
In the particular case of GZSL, as the results may be biased toward seen classes,
a common metric is to report the Harmonic mean (HM) of the
the measures for seen and unseen classes (whether Acc or mIoU).

\subsection{Backbone feature extractors and generators}

\textbf{Feature extractors.}
Many backbones allow point cloud classification and segmentation. We experimented with 4, illustrating our method is not backbone-dependent. For classification, we chose PointNet as in  \cite{cheraghian2019mitigating,cheraghian2019zero} to enable comparisons and to show our results are not just due to a better backbone. Besides, PointNet is only 4\% less accurate than the state of the art on ModelNet40. For segmentation, in the absence of prior work, we chose three backbones at the state of the art when trained under full supervision: ConvPoint~\cite{boulch2020convpoint} for S3DIS, FKAConv~\cite{boulch2020fkaconv} for ScanNet, and KPConv~\cite{thomas2019kpconv} for SemanticKITTI. Each backbone is trained on the seen classes using the recommended setting described in its respective paper.

\textbf{Generators.}
We tested 4 generators, as in \cite{Bucher_2017_ICCV,bucher2019zero}: a denoising auto-encoder (DAE) \cite{bengio2013generalized}, a generative moment-matching network (GMMN)~\cite{li2015generative}, a conditional GAN~\cite{odena2017conditional} and an adversarial auto-encoder~\cite{makhzani2015adversarial}.
We chose the best performing generator on the validation sets: DAE for classification and GMMN for segmentation (see supp.\,mat.).

\subsection{Parameter setup}

Parameters governing the training are selected by cross-validation. We create validation sets out of the training sets.

\textbf{Validation sets.}
To create cross-validation splits, we follow the ZSL protocol of \cite{Bucher_2017_ICCV}: we randomly select $20\%$ or at least 2 of the seen classes of the training data as validation classes. The feature backbone and the generator are trained from scratch on each split, using only the seen classes not selected for validation. 
Splits are evaluated only on validation classes.
For semantic segmentation, some classes are present in almost every pillar (chunk to process for point clouds), e.g., floor or ceiling for indoor scenes; they cannot be selected as unseen validation class.

\begin{figure}
    \centering
    \includegraphics[width=\linewidth]{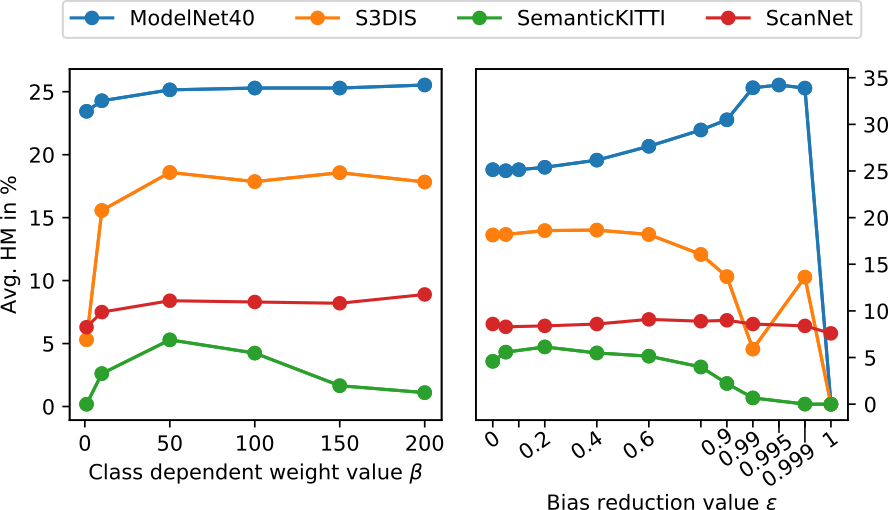}
    \caption{Effect of bias reduction parameters on HmAcc (ModelNet40 classification) and HmIoU (semantic segmentation on S3DIS, ScanNet, SemanticKITTI): (a) class-dependent weighting factor $\beta$, (b) calibrated stacking value $\epsilon$ (average over 20 runs).}
    \label{fig:exp:bias}
\end{figure}

\textbf{Number of generations.}
In ZSL, the final classifier $f(\cdot)$ is trained on artificial examples.
For classification, a study of the impact of the number of generated examples at training allows to observe that: first, a very small dataset does not perform well and second, a plateau is reached at about 100 samples per class, with a maximum around 500, which is the value we select.
For semantic segmentation, we follow the procedure from~\cite{bucher2019zero}, where object representations are generated according to their frequency in the dataset.

\textbf{Bias reduction in GZSL.}
As $\beta$ is a training parameter and $\epsilon$ is used for post-processing, we tune them sequentially with cross-validation. A range study is shown on Fig.\,\ref{fig:exp:bias}.

\emph{Class-dependent weighting.}
We observe on Fig.\,\ref{fig:exp:bias}(a) the relative stability of the evaluation metric on all datasets/ tasks when $10<\beta<100$. As a close-to-maximum value is reached for $\beta=50$, we use this value in all experiments.

\emph{Calibrated stacking.}
We then select a value of $\epsilon$. 
As the best $\epsilon$ varies substantially for each dataset (see Fig.\,\ref{fig:exp:bias}(b)), we choose it dataset-wise: $0.995$ for ModelNet40, $0.4$ for S3DIS, $0.6$ for ScanNet and $0.2$ for SemanticKITTI.

Note that class-dependent weighting ($\beta\,{>}\,1$) yields a gain up to 2 pts on classification (see Fig.\,\ref{fig:exp:bias}(a), compared to $\beta\,{=}\,1$) while calibrated stacking gains up to 8 pts. However, it brings a gain up to 13 pts on the segmentation task, which fully justifies the use of both bias reduction techniques.

\subsection{Benchmark results}

We now present the results of our method on the benchmarks.
As opposed to the previous section, where parameter studies were made on small validation sets created out of the training sets, we train here on the whole training sets of seen classes, which explains the better results we obtain.

\begin{table*}
\centering
\begingroup
\renewcommand{\arraystretch}{0.9}
\setlength{\tabcolsep}{1.8pt}
\begin{tabular}
{l|c|c|c|c|c|c|ccc|ccc|ccc}
\toprule
\multicolumn{2}{c|}{} & \multirow{3}{*}{\makecell{Full\\super-\\vision}} &\multicolumn{3}{c|}{ZSL} & \multicolumn{10}{c}{GZSL}\\
\cmidrule{4-16}
\multicolumn{2}{c|}{Method} & & \multirow{2}{*}{W2V} & \multirow{2}{*}{GloVe} & Glove& &\multicolumn{3}{c|}{W2V} & \multicolumn{3}{c|}{GloVe} & \multicolumn{3}{c}{GloVe + W2V}\\ 
 & Gener- & & & & + W2V & Bias & Acc. & Acc. & \multirow{2}{*}{HM} & Acc. &Acc. & \multirow{2}{*}{HM} & Acc.& Acc. & \multirow{2}{*}{HM}\\
& ative  & Acc. & Acc. & Acc. & Acc.  & reduct. & $\seen$ & $\unseen$ &  & $\seen$ & $\unseen$ & & $\seen$ & $\unseen$ & \\ 
\midrule
\rowcolor{LightGrey}
\multicolumn{2}{l|}{PointNet~\cite{qi2017pointnet}} & 89.2 & \multicolumn{13}{c}{}\\
\midrule
f-CLSWGAN* \cite{xian2018feature} & \checkmark  &\cellcolor{LightGrey}&20.7 & - & - &  & 76.3 & 3.7 & 7.0 & - & - & - & - & - & -  \\
CADA-VAE*  \cite{schonfeld2019generalized}  & \checkmark   &\cellcolor{LightGrey}&23.0 & - & - &  & 84.7 & 1.3 & 2.6 & - & - & - & - & - & -  \\
ZSLPC \cite{cheraghian2019zero} &   &\cellcolor{LightGrey}& 28.0 & 20.9 & 20.5 &  & 40.1 & 22.5 & 28.8 & 49.2 & 18.2 & 26.6 & - & - & - \\
MHPC \cite{cheraghian2019mitigating} &  &\cellcolor{LightGrey}& \textbf{33.9} & 28.7 & -  & \checkmark  & \textbf{53.8} & 26.2 & 35.2 & \textbf{53.8} & 25.7 & \textbf{34.8} & - & - & -\\
3DGenZ (ours) & \checkmark & \cellcolor{LightGrey}& 28.6 & \textbf{29.3} & \textbf{36.8} & \checkmark  & 48.8	& \textbf{29.3} & \textbf{36.6} & 44.7 & \textbf{28.4} & 34.7 & \textbf{47.8} & \textbf{36.5} &\textbf{41.3} \\
\bottomrule
\multicolumn{1}{l}{}\\[-7pt]
\end{tabular}
\endgroup
\caption{ZSL and GZSL classification results (in \%) on ModelNet40. *: adaptation of 2D methods to 3D point clouds, implemented in~\cite{cheraghian2020transductive}. Results for ZSLPC are with the best reported variant, i.e., PointNet + NetVlad \cite{cheraghian2019zero}. For fair comparison we report results with the same PointNet backbone. Results are averaged over 20 runs for our method.}
\vspace*{3mm}
\label{tab:zsl_gzsl_overview}
\end{table*}

\begin{table*}[t]
\renewcommand{\arraystretch}{0.9}
\setlength{\tabcolsep}{3pt}
\centering
\begin{tabular}{l|c|c||rrr|>{\columncolor{HL}}r||rrr|>{\columncolor{HL}}r||rrr|>{\columncolor{HL}}r}
\toprule
\multicolumn{1}{c}{}& \multicolumn{2}{c||}{Training set} & \multicolumn{4}{c||}{S3DIS} & \multicolumn{4}{c||}{ScanNet} &  \multicolumn{4}{c}{SemanticKITTI} \\
& Back- & Classi- &
\multicolumn{3}{c|}{mIoU} & \multicolumn{1}{c||}{\cellcolor{HL}{HmIoU}} & 
\multicolumn{3}{c|}{mIoU} & \multicolumn{1}{c||}{\cellcolor{HL}{HmIoU}} & 
\multicolumn{3}{c|}{mIoU} & \multicolumn{1}{c}{\cellcolor{HL}{HmIoU}} \\
&  bone & fier  & 
\multicolumn{1}{c}{$\seen$} & \multicolumn{1}{c}{$\unseen$} & All & \cellcolor{white}{} &
\multicolumn{1}{c}{$\seen$} & \multicolumn{1}{c}{$\unseen$} & All & \cellcolor{white}{}  &
\multicolumn{1}{c}{$\seen$} & \multicolumn{1}{c}{$\unseen$} & All & \cellcolor{white}{}  \\
\midrule
\rowcolor{LightGrey}
\multicolumn{15}{l}{\textit{Supervised methods with different levels of supervision}}\\
\rowcolor{LightGrey}
Full supervision & $\seen \cup \unseen$ & $\seen \cup \unseen$ & 74.0 & 50.0  & 66.6 &59.6~~ & 43.3&51.9&45.1&47.2& 59.4 & 50.3 & 57.5 & 54.5~~  \\
\rowcolor{LightGrey}
ZSL backbone & $\seen$ & $\seen \cup \unseen$   & 60.9 & 21.5 & 48.7 & 31.8~~ &41.5&39.2&40.3& 40.3& 52.9 & 13.2& 42.3 & 21.2~~  \\
\rowcolor{LightGrey}
ZSL-trivial & $\seen$ & $\seen$ & 70.2 & 0.0   & 48.6 & 0.0~~ &39.2  &0.0&31.3&0.0& 55.8 & 0.0 & 44.0 & 0.0~~  \\
\midrule
\multicolumn{15}{l}{\textit{Generalized zero-shot-learning methods}}\\
\rowcolor{lightred}
ZSLPC-Seg*~\cite{cheraghian2019zero}$^\dagger$ & $\seen$ & $\unseen$  & 65.5 & 0.0 & 45.3& 0.0~~  &28.2 &0.0&22.6&0.0& 49.1 & 0.0 & 34.8& 0.0~~\\
\rowcolor{lightred}
DeViSe-3DSeg*~\cite{frome2013devise}$^\dagger$ & $\seen$ & $\unseen$   & 70.2 & 0.0 & 48.6 & 0.0~~  & 20.0&0.0&16.0&0.0& 49.7 & 0.0 & 36.6& 0.0~~\\ 
ZSLPC-Seg~\cite{cheraghian2019zero}$^\dagger$ & $\seen$ & $\unseen$  & 5.2 & 1.3 & 4.0& 2.1~~  & 16.4&4.2&13.9&6.7& 26.4 & 10.2 & 21.8& 14.7~~\\
DeViSe-3DSeg~\cite{frome2013devise}$^\dagger$ & $\seen$ & $\unseen$   & 3.6 & 1.4 & 3.0& 2.0~~  &12.8 &3.0&10.9&4.8& 42.9 & 4.2 & 27.6& 7.5~~\\
3DGenZ (ours) & $\seen$ & $\seen \cup \hat{\unseen}$  & \textbf{53.1} & \textbf{7.3}&  \textbf{39.0} & \textbf{12.9}~~ & \textbf{32.8} & \textbf{7.7} & \textbf{27.8} & \textbf{12.5} & \textbf{41.4} & \textbf{10.8} & \textbf{35.0} & \textbf{17.1}~~\\
\bottomrule
\multicolumn{1}{l}{}\\[-7pt]
\end{tabular}
\caption{GZSL semantic segmentation results (in \%).
$^\dagger$Our adaption of the method. *Direct, unrepaired (failing) adaptation.}
\label{tab:sem_seg_overview}
\end{table*}

\begin{figure}[t]
    \centering
    \includegraphics[width=\linewidth]{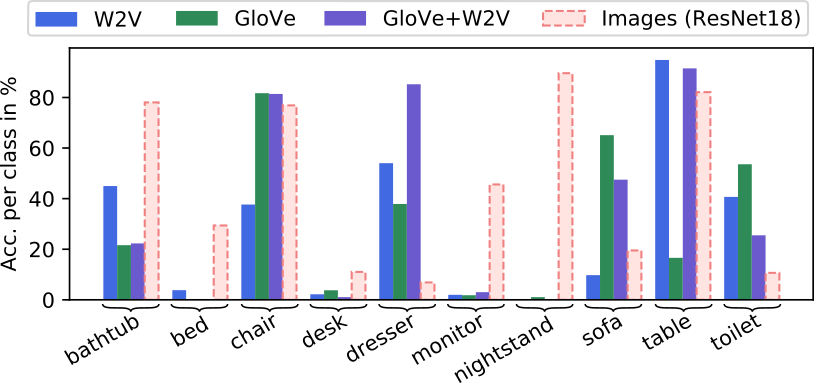}
    \caption{Per-class accuracy for each unseen class of ModelNet40, for each kind of embedding (word and image, see Sect.\,\ref{sec:im2pc}).}
    \label{fig:cls_zsl_embeddings}
\end{figure}

\subsubsection{Classification}

We evaluate classification on ModelNet40 \cite{wu20153d} using classes of ModelNet10 as unseen, like \cite{cheraghian2019zero}, to allow comparison.
Classification results for both ZSL and GZSL are presented in Table~\ref{tab:zsl_gzsl_overview} (averaged over 20 runs for our method).

\textbf{Influence of auxiliary information.}
Results are presented for three class prototypes: Word2Vec (W2V)~\cite{mikolov2013distributed}, GloVe~\cite{pennington2014glove}, and their concatenation (GloVe+W2V).
For both ZSL and GZSL, we observe that, while using W2V or GloVe alone leads to similar performances, their concatenation performs much better: $+7.5\%$ (ZSL), $+5.4\%$ (GZSL). 

Fig.\,\ref{fig:cls_zsl_embeddings} provides detailed ZSL accuracy for each unseen class and each kind of class prototype (see supp.\ mat.\ for GZSL).
We see that some classes (\emph{bed}, \emph{desk}, \emph{monitor}, \emph{nightstand}) are incorrectly predicted (if predicted at all) regardless of the class prototype.
Except for \emph{bathtub} and \emph{toilet}, W2V+GloVe outperforms the worst of the two embeddings by a large margin. No performance gain is however guaranteed by using W2V+GloVe over W2V or GloVe alone.

\textbf{Comparison with state-of-the-art methods.}
Tab.\,\ref{tab:zsl_gzsl_overview} reports scores for ZSLPC \cite{cheraghian2019zero} and MHPC \cite{cheraghian2019mitigating} as well as the adaptation to 3D point clouds of the 2D methods f-CLSWGAN* \cite{xian2018feature} and CADA-VAE* \cite{schonfeld2019generalized} proposed in~\cite{cheraghian2020transductive}.
All methods, including ours, use the same backbone.

For ZSL, our method performs the best with GloVe and places second with W2V.
Interestingly, it establishes the new state of the art with W2V+GloVe, while the previous state-of-the-art method \cite{cheraghian2019zero} shows a significantly lower accuracy when combining both embeddings.
It could be due to the nearest-neighbor search in a higher dimension space.

For GZSL, our approach is less accurate than baselines on seen classes. It is due to the parameter setting policy of our bias reduction techniques, that we set to favor similar scores for seen and unseen classes, trading seen-class accuracy for more accurate unseen classes. The fact is we outperform on HmAcc the other methods for W2V, and reach the state of the art for GloVe. With W2V+GloVe, we outperform the state of the art by a significant margin.

This shows that, contrary to what was believed \cite{cheraghian2020transductive}, 2D generative ZSL methods can successfully be transferred to 3D, and even outperform non-generative ZSL methods on point clouds. In particular, we invalidate the hypothesis that the success of 2D generative models relies on pre-trained models \cite{cheraghian2020transductive}:
all our networks are trained from scratch, only on seen classes of moderately-sized 3D datasets.
Transfer to 3D however is not straightforward. In particular, contrary to a number of other 2D ZSL methods, 2D generative approaches are known not to require reducing a bias towards seen classes, as they work at feature level and can generate as many unseen examples as needed. Yet, because 3D backbones trained from scratch are somehow too specialized on seen classes, compared to 2D backbones pre-trained on huge datasets, we had to resort to two bias reduction techniques to outperform the state of the art on 3D ZSL.

\begin{figure*}[!t]
    \centering
    \includegraphics[width=\linewidth]{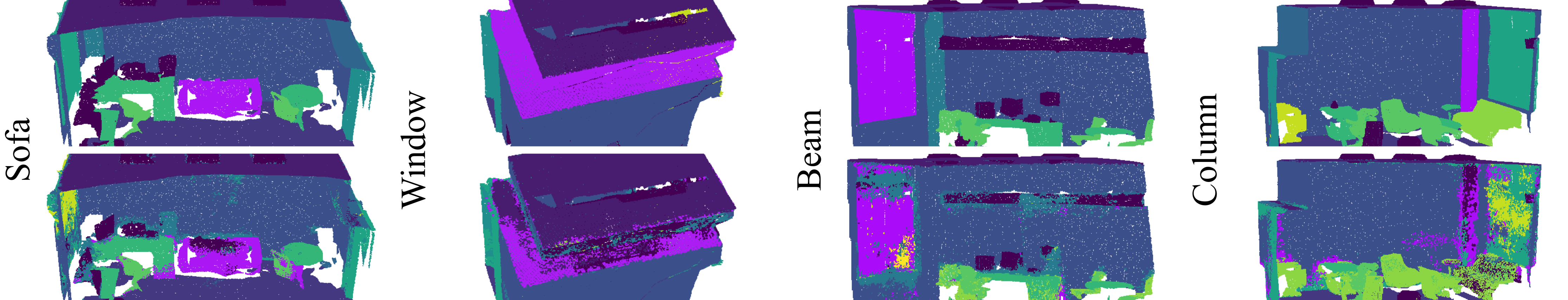}
    
    \vspace{2pt}
    \includegraphics[width=\linewidth]{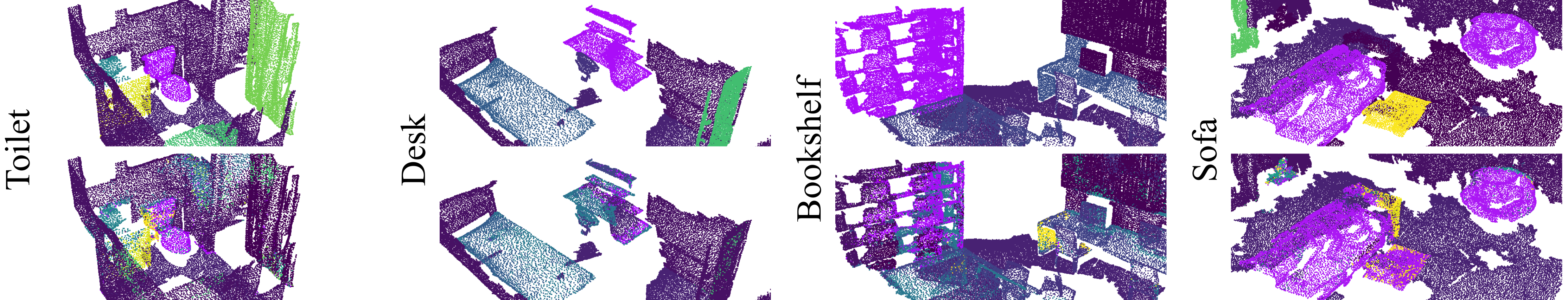}
    
    \vspace{2pt}
    \includegraphics[width=\linewidth]{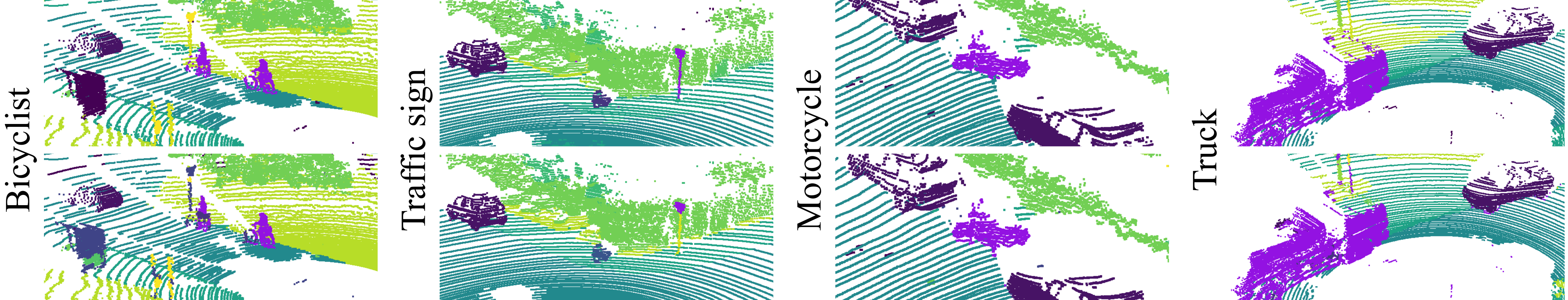}
    
    \caption{Zero-shot segmentation on scenes from S3DIS (row 1-2), ScanNet (row 3-4) and SemanticKITTI (row 5-6). For each two-row block, first row is the ground truth, second row is our ZSL prediction. Unseen classes have pink colors.}
    \vspace*{-1.5mm}
    \label{fig:exp:visu_sk}
\end{figure*}

\subsubsection{Semantic Segmentation}

We now present results for semantic segmentation on S3DIS, ScanNet and SemanticKITTI.
As we are the first to address this task, there is no method to compare with. We however define two baselines, before comparing them to our method.
The scores are reported in Table~\ref{tab:sem_seg_overview}.

\textbf{Supervision as upper bound.}
To scale our results, we train three supervised models (see grey rows in Tab.\,\ref{tab:sem_seg_overview}): with the backbone feature extractor and the classifier trained on all classes, seen and unseen (full supervision); with the backbone trained on seen classes only, and the classifier on all classes (ZSL backbone); with the backbone and classifier trained on seen classes only (ZSL-trivial).
The first model is an upper bound when all classes are seen; the second provides an upper bound for zero-shot methods.
It performs noticeably worse than the fully-supervised model, which hints that the backbone feature extractor trained only on seen classes generates object representations for unseen classes that are not easily distinguishable from seen classes.

For seen classes, our method behaves similarly with all datasets: it reaches a mIoU around 10 points below the maximum score (reached by ZSL backbone and a fully-supervised classifier). However, we notice different a behaviour on the unseen classes. While on SemanticKITTI, our performance is close to the performance of the ZSL backbone, we observe a much larger gap on S3DIS and ScanNet. This suggests that the generation of class prototypes for the unseen classes is of lower quality on S3DIS and ScanNet than on SemanticKITTI.

\textbf{Baselines.}
We create two baselines for zero-shot 3D segmentation: (1)~ZSLPC-Seg is an adaptation of the ZSL classification method ZSLPC~\cite{cheraghian2019zero} to segmentation.  ZSLPC is itself an adaptation of ConSe~\cite{norouzi2013zero} from 2D images to 3D point clouds. (2)~DeViSe-3DSeg is an adaptation of Devise-Seg to 3D point clouds, Devise-Seg being itself an adaptation of DeViSe \cite{frome2013devise} from classification to segmentation, as proposed in \cite{bucher2019zero}.
The two baselines rely on a nearest-neighbor search, either in the class prototype space or in the object representation space.
However, experimentally, searching the $K\,{=}\,1$-nearest neighbors leads to no prediction at all for unseen classes (light-red rows in Table~\ref{tab:sem_seg_overview}). It illustrates that adapting a ZSL method to 3D or to segmentation is not straightforward.
To construct meaningful baselines, we modify these methods by looking for the nearest unseen class among the $K$-nearest neighbors. If no such unseen class is found, then the class of the closest neighbor is selected.
The values of $K$ that maximize the HmIoU are:
for DeViSe-3DSeg, $K\,{=}\,7, 2, 5$ on S3DIS, ScanNet, SemanticKITTI, respectively; for ZSLPC-Seg, $K\,{=}\,5, 2, 5$.

\textbf{Results.}
Our approach outperforms the baselines by large margins and establishes the state of the art on the three defined benchmark. According to the HmIoU metric, the gap is much larger on S3DIS than on SemanticKITTI and ScanNet.
It is possible that our method deals relatively better with a smaller number of classes, while the baseline methods benefit more from a larger number of classes.

Figure~\ref{fig:exp:visu_sk} provides qualitative results of our segmentation method on the three datasets.
On S3DIS, the object are correctly located, but the classifier hesitates between classes, e.g., between window and board. As a result, predictions are mixed, which produces a salt-and-pepper effect.
On ScanNet, the unseen classes \emph{sofa} and \emph{toilet} seem well segmented, with a bit of oversegmentation on \emph{sofa}. On the contrary, the network has more difficulties to segment \emph{desks} and \emph{bookshelves}.
On SemanticKITTI, objects appear much better segmented even though the mIoU for unseen classes in Table~\ref{tab:sem_seg_overview} is close to the one on S3DIS.
The network is more confident, resulting in consistent segmentations, but it oversegments some classes.
The semantic similarity between a pole and a traffic-sign, and between a two-wheeler with or without a rider, is a useful cue for ZSL, but an issue for GZSL, that has to tell them apart. In fact, our bias reduction boosts traffic signs and bicyclists so much that poles and unridden bicycles are not segmented anymore. Yet, unseen trucks are segmented without much altering seen vehicle segmentation, except for \emph{other vehicles}.

It is remarkable that our framework proves to work both for classification and segmentation. Not all classification models adapt well to semantic segmentation. In fact, the two baselines DeViSe-3DSeg and ZSLPC-Seg originating from ZSL classification methods \cite{cheraghian2019zero, frome2013devise} do not transfer well to semantic segmentation, despite our improvements.

\subsection{Image-based 3D zero-shot learning}
\label{sec:im2pc}
The auxiliary information for ZSL provides descriptions of unseen classes.
Our method uses W2V and GloVe representations as auxiliary information, which are created from word co-occurrences in text corpora.
We have shown in the previous sections that it is possible to meaningfully link these class prototypes to the point cloud representations.

We propose to investigate the use of an alternative auxiliary information based on image representations.
As images capture the appearance of objects, visual representations should better link to object geometry.
Here we describe an object class with a small set of images, and generate a visual representation by averaging features extracted using a CNN pre-trained on large datasets such as ImageNet~\cite{russakovsky2015imagenet}.
Although rare, image-based representations have already been used for ZSL, e.g., human action recognition in videos \cite{wang2017alternative}.

We experiment with image representations as class prototype for 3D ZSL and GZSL.
We consider 2 different image encoders, one pre-trained with supervision (ResNet-18~\cite{he2016deep}) and one with self-supervision (ResNet-50~\cite{caron2020unsupervised}).
Experiments are run with the same parameters as for text embeddings.
To generate prototypes for each class (seen and unseen), we average the features of the top-100 images obtained by a Google search with the class name.
Results are shown in Table~\ref{tab:image_emb}. Results for word embeddings are recalled from Tables~\ref{tab:zsl_gzsl_overview} and~\ref{tab:sem_seg_overview}, for comparison purposes.

\begin{table}
    \centering
    \renewcommand{\arraystretch}{0.9}
    \setlength{\tabcolsep}{2pt}
    \begin{tabular}{l|cc|cc}
    \toprule
                        & \multicolumn{2}{c|}{\emph{Classif.}} & \multicolumn{2}{c}{\emph{Segmentation}} \\
                    & \multicolumn{2}{c|}{ModelNet40} & ScanNet & KITTI \\
    Representation  &  ZSL   & GZSL & \multicolumn{2}{c}{HmIoU}\\
    \midrule
    \rowcolor{LightGrey}
    W2V+GloVe (self-sup.) & 36.8 & \textbf{41.3} & 12.5 & \textbf{17.1}\\
    \midrule
    ResNet-18~\cite{he2016deep} (sup.)        & \textbf{43.6}  & 40.0 & 13.9 & 3.6
    \\
    ResNet-50~\cite{caron2020unsupervised} (self-sup.) & 37.0  & 36.5 & \textbf{15.5} &5.3\\
    \bottomrule
    \multicolumn{1}{l}{}\\[-7pt]
    \end{tabular}
    \caption{Image-based 3D G/ZSL classification and segmentation.}
    \label{tab:image_emb}
    \vspace*{-2mm}
\end{table}

For ZSL and GZSL, the two backbones performs as well or better than their counterpart for text embeddings, even without further tuning of the $\epsilon$ and $\beta$ parameters (GZSL case).
Image embeddings even outperform text embedding for segmentation on ScanNet, but fail on SemanticKITTI.
A reason to this failure may be the presence of multiple classes per images (e.g. bicycle/ist, motorcycle/ist, sign/pole) leading to indistinguishable representations.

SemanticKITTI left aside, the good performance of the self-supervised representations underlines that this image-based approach can be a competitive alternative to the usual text-based ZSL. Indeed, in the same spirit as the non-supervised word embeddings of W2V, our image-based approach requires little or no supervision: merely the manual collection of a small quantity of images associated to given class names. Moreover, looking at the class-wise accuracy (for ResNet-18) on Fig.\,\ref{fig:cls_zsl_embeddings}, we observe that the distribution of accuracies over the classes with an image-based representation is significantly different than with the text-based representations. In particular, classes like \emph{nightstand} and \emph{monitor} perform much better with image-based representations as auxiliary data. Taking the best of the two embeddings is a promising perspective and is left as future work.

\section{Conclusion}
In this study, we present a generative method for zero-shot learning on 3D point clouds.
Experiments on a classification task shows that our method reaches the state of the art in both a classical and a generalized setting. Additionally, we show that our method can be easily extended to zero-shot semantic segmentation.
To our knowledge, we are the first to tackle this task.
We define natural baselines for 3D zero-shot segmentation, based on state-of-the-art approaches for classification, and compare them to our approach.
Our method outperforms them on the three indoor and outdoor datasets we propose, based on S3DIS, ScanNet and SemanticKITTI. Besides, we introduce the use of image-based representations as an alternative auxiliary data for 3D ZSL and GZSL.
We show that it is possible to outperform text-based representations in ZSL for classification. This experiments opens new perspectives as we observe that text embedding and image embeddings produce different performance distribution. Future work includes merging the two types of information to maximize zero-shot efficiency, as well as using phrasal (multi-word) embeddings to discover complex corner cases in large datasets.

\ifarxiv
\begin{figure*}[t]
    \centering
\Large \bf Generative Zero-Shot Learning for Semantic Segmentation of 3D Point Clouds \\ --- Supplementary Material ---
\end{figure*}
\appendix

\setcounter{figure}{5}
\setcounter{table}{3}

We present here complementary information on the 3DV 2021 paper ``Generative Zero-Shot Learning for Semantic Segmentation of 3D Point Clouds''.
It provides more details and results on the following topics:

\begin{itemize}[topsep=3pt,itemsep=-1pt]
    \item[\ref{sup:sec:generator}.] Generators: 
    \begin{enumerate}[nosep,topsep=-4pt]
        \item implementation,
        \item comparison,
        \item number of generated representations.
    \end{enumerate}
    \item[\ref{sup:sec:classification}.] Classification:
    \begin{enumerate}[nosep,topsep=-4pt]
        \item validation splits on ModelNet40,
        \item standard deviation, best and worst accuracy,
        \item sensitivity to the number of seen classes,
        \item ZSL confusion matrices,
        \item GZSL confusion matrices,
        \item classification results on McGill and SHREC2015\rlap.
        \item ablation study of bias reduction,
    \end{enumerate}
    \item[\ref{sup:sec:semantic_segmentation}.] Semantic segmentation:
    \begin{enumerate}[nosep,topsep=-4pt]
        \item cross-validation splits,
        \item semantic segmentation baselines,
        \item classwise performance,
        \item alternative splits,
        \item visualization of class prototype spaces,
        \item upper bound for semantic segmentation.
    \end{enumerate}
    \item[\ref{sup:sec:image_representation}.] GZSL with image-based embeddings:
    \begin{enumerate}[nosep,topsep=-4pt]
        \item image selection,
        \item feature averaging and normalization,
        \item classwise performance,
        \item visualization of class prototype spaces,
        \item sensitivity to the image collection quality.
    \end{enumerate}
    
\end{itemize}

After some cleaning, code should be available in the fall of 2021 on \url{https://github.com/valeoai}. Stay tuned\rlap{.}

\section{Generators}
\label{sup:sec:generator}

We experimented with 4 different kinds of generators (see Section~4.2 of the paper): 
\begin{itemize}[nosep]
    \item a denoising auto-encoder (DAE) \cite{bengio2013generalized},
    \item a generative moment-matching network (GMMN) \cite{li2015generative}\rlap,
    \item a conditional GAN (AC-GAN)~\cite{odena2017conditional},
    \item an adversial auto-encoder (AAE)~\cite{makhzani2015adversarial}.
\end{itemize}
We detail here their implementation and compare them.

\ifarxiv
\setcounter{figure}{1}
\begin{figure}
    \centering
    \includegraphics[width=\linewidth]{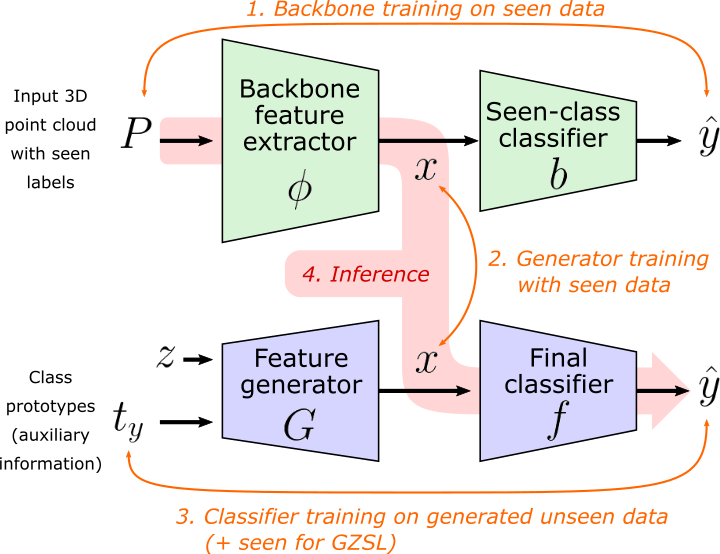}
    \caption{(reminder) Training and inference: (1)~backbone training on the seen classes, (2)~generator training, (3)~classifier training with artificial unseen features for ZSL (unseen and seen for GZSL), (4)~inference through backbone and final classifier.}
\end{figure}
\setcounter{figure}{5}
\fi

\subsection{Implementation details}

To implement the generators, we basically follow the settings described in more details in \cite{Bucher_2017_ICCV,bucher2019zero}.

Each of these generators is made of 2 fully-connected layers, although with a different architecture.
\begin{itemize}[topsep=3pt,itemsep=-1pt]
    \item The DAE consists of a encoder and a conditional decoder; it is trained with a mean square error loss.
    \item The AAE extends the DAE with a discriminator to constrain the latent code with an adversial criterion so that it follows a (normal) prior distribution; compared to the DAE, the AAE loss include an additional adversarial loss term. 
    \item The AC-GAN generator is trained to produce conditional distributions similar to the true distributions; the training loss is the sum of a multi-label cross-entropy loss and an adversarial binary cross-entropy loss. 
    \item Last, the GMMN is a conditional MLP trained with a loss penalizing the maximum mean discrepancy.
\end{itemize}

The number and size of the layers is kept consistent with \cite{Bucher_2017_ICCV, bucher2019zero}; we did not try to optimize them. The experiments show that this setting also works well for generating 3D-like features from textual class prototypes as well as from 2D image features.
More generally, we believe that generative ZSL methods developed for 2D images can be transferred well in this manner to handle 3D point clouds.

For training, we use the Adam optimizer~\cite{kingma2014adam} as in \cite{Bucher_2017_ICCV, bucher2019zero}, keeping the same parameters regarding the learning rate (decay) and the number of training epochs. As we use different datasets, the definition of one training epoch is however a bit different. 
\begin{itemize}[topsep=3pt,itemsep=-1pt]
    \item For classification, we show every point cloud of the training set once in every epoch.
    \item For semantic segmentation, we use the default definition of an epoch for the respective backbone \cite{boulch2020convpoint, thomas2019kpconv}.
\end{itemize}

\begin{table*}
\centering
\begin{tabular}{llll|cccc}
\toprule
Dataset & Task& Setting & Metric &  DAE & GMMN & AC-GAN & AAE\\
\midrule
ModelNet40 & classif. & ZSL & \multicolumn{1}{l|}{\% HM} & \textbf{36.0} & 30.5  & 34.6 & 29.8 \\
S3DIS & segment. & GZSL & \% HmIoU & 15.9 & \textbf{18.6}  & 10.3 & 12.4\\
ScanNet & segment. & GZSL & \% HmIoU & 7.2 & \textbf{9.1}  & 6.5 & 6.3\\
SemanticKITTI & segment. & GZSL & \% HmIoU & 4.4 & \textbf{6.1} & 4.1 &4.3\\
\bottomrule
\\[-7pt]
\end{tabular}
\caption{Comparison of the performance of the different generators for classification (on ModelNet40) and for semantic segmentation (on S3DIS, ScanNet and SemanticKITTI), based on the validation splits. The best generators regarding this validation data (DAE and GMMN) are kept for testing.}
\label{tab:exp:generator}
\end{table*}

\subsection{Generator comparison}

We ran cross-validation experiments with the four kinds of generators.
We studied the behavior of the generators on ModelNet40 for classification and on S3DIS, ScanNet and SemanticKITTI for semantic segmentation.  

For each dataset/task, we only used validation data created out of the available bases classes. The construction of validation sets is described in Section~\ref{sec:valid_split_classif} for classification, and Section~\ref{sec:cross_val_seg} for semantic segmentation.
Results are presented in Table~\ref{tab:exp:generator}.
We remark that the generators do not perform similarly and, more interestingly, that the best generator vary from one dataset/task to the other.
This is an observation also made in~\cite{Bucher_2017_ICCV}.

\subsection{Number of generated representations}

With the cross-validation splits for ModelNet40, we study the impact of the number of generated examples for classification, as shown in Figure~\ref{fig:exp:nb_images_hyper}. The maximum top-1 accuracy is reached with 500 generated representations. We then use this number of generations for the (G)ZSL classification task.

\begin{figure}[t]
    \centering
    \includegraphics[width=\linewidth]{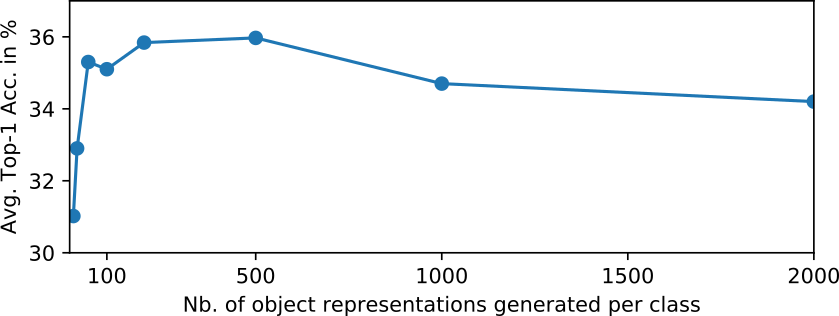} 
    \caption{Cross-validation Acc (5-split average) 
    wrt.\ the number of generated examples per class in ZSL classification (ModelNet40).}
    \label{fig:exp:nb_images_hyper}
\end{figure}

Based on these experiments, we chose to use DAE for classification tests on ModelNet40 and GMMN for semantic segmentation tests on S3DIS, ScanNet and SemanticKITTI.

\section{Classification}
\label{sup:sec:classification}

\subsection{Validation splits on ModelNet40}
\label{sec:valid_split_classif}

Cross-validation is done with 5 splits on
ModelNet40, i.e., on the 30 (seen) classes not in the 10 (unseen) classes of ModelNet10. Following \cite{Bucher_2017_ICCV}, we select as validation classes in each split 20\% of the seen classes, i.e., 6 classes.

\begin{table}[!t]
\centering
\setlength{\tabcolsep}{0.1cm}
\def\stdsize#1{\small \textcolor{MidnightBlue}{#1}}
\begin{tabular}{l|rrr|rrr}
\toprule
Setting & \multicolumn{3}{c|}{ZSL} & \multicolumn{3}{c}{GZSL}\\
\midrule
Metric & \multicolumn{3}{c|}{Top-1 Acc (\%)} & \multicolumn{3}{c}{Acc-HM (\%)}\\
\midrule
Word & W2V & GloVe & GloVe& W2V & GloVe & GloVe\\ 
embedding & & & +W2V & & &+W2V \\
\midrule
\stdsize{Worst} &\stdsize{25.5} &\stdsize{23.5} &\stdsize{33.3} &\stdsize{33.5}
&\stdsize{31.8}
&\stdsize{36.4}\\
Average  & 28.6 & 29.3 & 36.8 & 36.6 & 34.7 & 41.3\\
\stdsize{Best}  &\stdsize{32.8}  &\stdsize{35.0} &\stdsize{39.7} &\stdsize{38.7} &\stdsize{38.1}
&\stdsize{44.6}\\
\stdsize{Std. deviation}  &\stdsize{2.1} &\stdsize{3.0}  &\stdsize{1.7} & \stdsize{1.6} 
&\stdsize{1.9} 
&\stdsize{2.1} \\
\bottomrule
\multicolumn{1}{l}{}\\[-7pt]
\end{tabular}
\caption{Variance study over 20 runs for zero-shot classification with our method on ModelNet40: ZSL Top-1 Acc (\%) and GZSL seen-unseen Acc-HM (\%), for different word embeddings.}
\label{supp:tab:cls_std_best_worst}
\end{table}

\begin{table*}[t]
\centering
\def\stdsize#1{\small \textcolor{MidnightBlue}{#1}}
\begin{tabular}{l|@{~~~}cccccccccc|c|c}
\toprule
& \!\!Bathtub\!\! & \!Bed\! & \!Chair\! & \!Desk\! & \!\!Dresser & \!\!\!\!Monitor\!\!\!\! & \begin{tabular}[c]{@{}c@{}}Night\! \\ stand\!\end{tabular} & \!Sofa\! & Table & Toilet  &\begin{tabular}[c]{@{}c@{}}Class \\ Acc.\end{tabular}&\begin{tabular}[c]{@{}c@{}}Global \\ Acc.\end{tabular} \\ \hline
W2V                                               & 45.0    & 3.8 & 37.7  & 2.2  & 54.0    & 2.0     & 0.1  & 9.8  & 94.8  & 40.7 & 29.0 & 28.6   \\ 
\stdsize{std deviation}    & \stdsize{14.8}    & \stdsize{3.7} & \stdsize{9.6}   & \stdsize{1.0}  & \stdsize{7.9}     & \stdsize{0.4}     & \stdsize{0.1}  & \stdsize{5.5}  & \stdsize{2.3}   & \stdsize{17.2}  & \stdsize{2.3} & \stdsize{2.1}  \\ \hline
GloVe                               & 21.6    & 0.0 & 81.7  & 3.8  & 37.9    & 1.8     & 1.1   & 65.1 & 16.6  & 53.6 & 28.3 & 29.3  \\ 
\stdsize{std deviation}    & \stdsize{7.4}     & \stdsize{0.0} & \stdsize{7.8}   & \stdsize{2.0}  & \stdsize{17.2}    & \stdsize{0.5}     & \stdsize{1.0}  & \stdsize{3.8}  & \stdsize{7.3}   & \stdsize{14.4}   &\stdsize{2.9} &\stdsize{3.0}\\ \hline
GloVe+W2V
& 22.3    & 0.1 & 81.3  & 1.1  & 85.2    & 3.0     & 0.1  & 47.5 & 91.5  & 25.5 & 35.8 & 36.8\\
\stdsize{std deviation}    & \stdsize{8.8}     & \stdsize{0.2} & \stdsize{7.3}   & \stdsize{1.0}  & \stdsize{2.3}     & \stdsize{0.5}     & \stdsize{0.3}  & \stdsize{6.3}  & \stdsize{6.3}   & \stdsize{11.0} & \stdsize{1.4} &\stdsize{1.7}\\
\bottomrule
\multicolumn{1}{l}{}\\[-7pt]
\end{tabular}
\caption{Classwise classification accuracy in the classical ZSL setting for W2V, GloVe and GloVe+W2V on the ModelNet40 benchmark, i.e., using the 10 classes of ModelNet10 as unseen classes.}
\label{supp:tab:cls_per_class}
\end{table*}

\begin{figure*}
\centering
\includegraphics[width=0.55\linewidth]{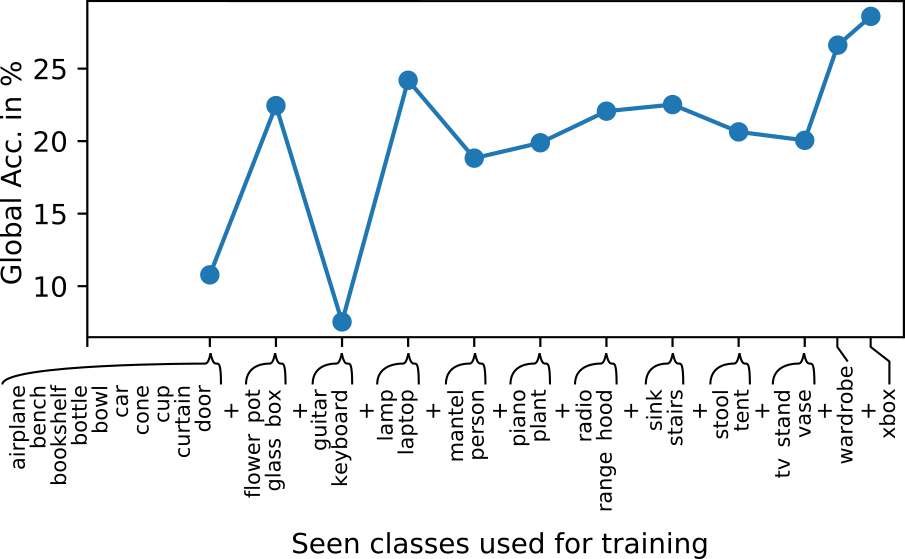}
\caption{Impact of the number of seen classes for training on the ZSL classification task.}
\label{supp:fig:alphabetic}
\end{figure*}

\subsection{Standard deviation, best and worst accuracy}

The classification results presented in the paper for our method are averaged over 20 runs.
We detail in Table~\ref{supp:tab:cls_std_best_worst} the standard deviations, as well as the results of the best and worst runs.
We consider that the observed standard deviations, between 1.6 and 3.0 points of accuracy, are understandable and acceptable given the difference of modality between the text-based class prototypes and the 3D features, that have to be bridged. Interestingly, combining GloVe and W2V not only leads to a better accuracy but also to a lower standard deviation for the ZSL task. However, in the GZSL setting, the standard deviation of combining Glove+W2V is a bit higher compared to using only Glove or W2V only.

More detailed results are shown in Table~\ref{supp:tab:cls_per_class}, where we report the average accuracy and standard deviation for each class, over 20 runs. 
Following \cite{cheraghian2019zero, cheraghian2019mitigating, cheraghian2020transductive, cheraghian2021zero}, we report the global accuracy (also known as the Top-1 Accuracy): a prediction is considered as correct if it matches the ground-truth class. Additionally, we also report the class accuracy (Acc.), which is the (unweighted) average of the classwise accuracy over all classes. As the ModelNet40 test set is relatively balanced, the difference  between the two kinds of average accuracies is small.

As noted in the paper, we observe that some classes like \class{bed}, \class{desk}, \class{monitor} and \class{nightstand} have a very low accuracy.  Unsurprisingly, the standard deviation of the accuracy for these classes is low as well.  More interestingly, the standard deviation tends to increase with the accuracy, except for class \class{table}, whose high accuracy is rather stable.  It is difficult to know the reason of this behavior without more advanced studies, that would however be quite specific to this dataset, given its moderate size and variety with respect to all involved parameters.  The causes of this high standard deviation may include: the absence of a similar-enough category among the seen classes; variations when training the generator, that never gets supervised information on unseen classes; weak correlations between textual embeddings and 3D features; and classification ambiguities.  Yet, when considering all classes together, both the average class accuracy and the global accuracy show a moderate standard deviation (around 2 points of accuracy).

\begin{figure*}[t]
\centering
\begin{tabular}[t]{ccc}
    \raisebox{-.5\height}{\includegraphics[width=0.28\linewidth]{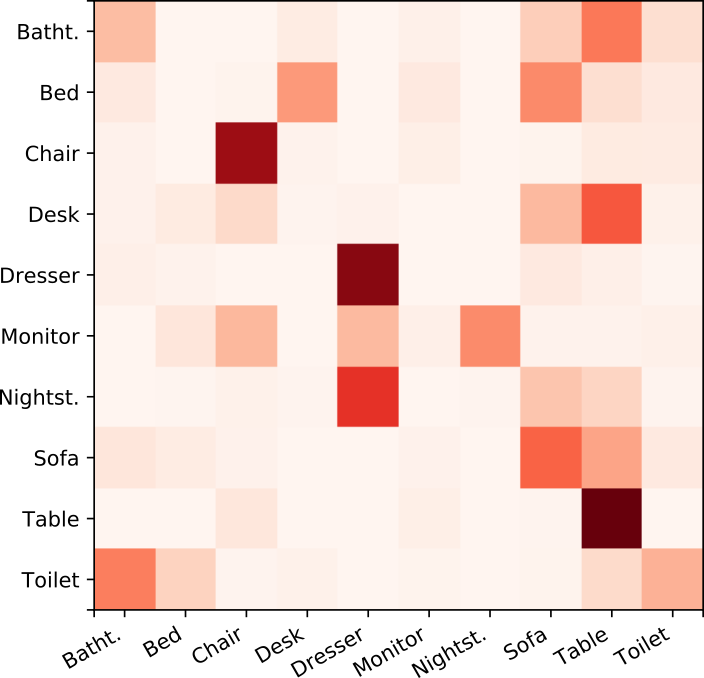}}&
    \raisebox{-.5\height}{\includegraphics[width=0.33\linewidth]{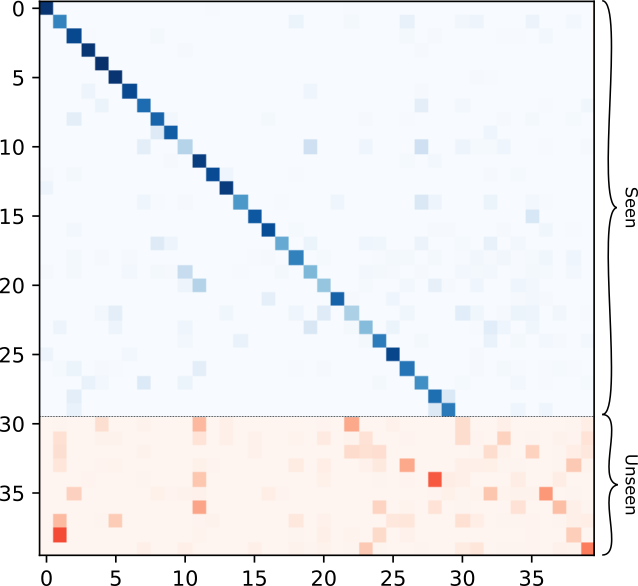}}&
    \raisebox{-.5\height}{\includegraphics[width=0.33\linewidth]{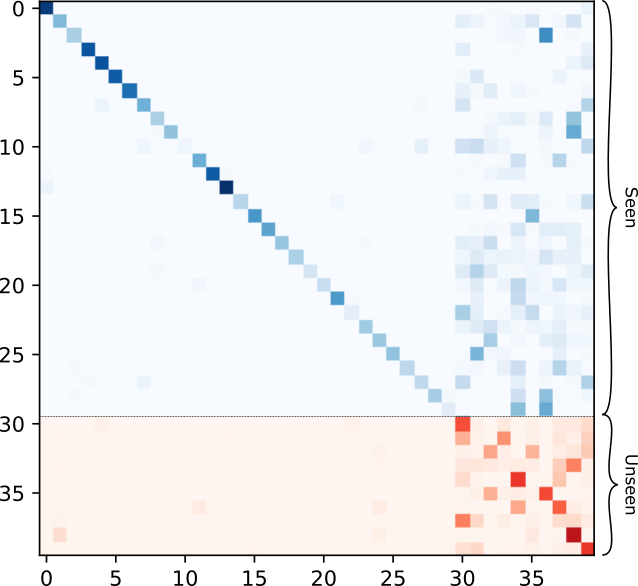}}\\
    (a) ZSL \raisebox{14pt}{}\raisebox{-8pt}{} & (b) GZSL with $\beta=50, \epsilon=0$ & (c) GZSL with $\beta=50, \epsilon=0.995$
\end{tabular}
\caption{Confusion matrices for zero-shot classifications in ZSL and GZSL settings.
The red color map is for unseen classes, and the blue color map for seen classes.
Each row shows the distribution of the predictions of one class: the darker the color is, the more often object points of this class are predicted as the class of the column.}
\label{supp:fig:cls_cm}
\end{figure*}

\subsection{Sensitivity to the number of seen classes}
\label{sup:sec:sensitivity}

To study the sensitivity of our method to the number of seen classes, we train our complete pipeline with a different number of seen classes on ModelNet40. Note that, for this analysis, we always evaluate the performance on the same 10 unseen classes of ModelNet40, but we train the networks using only on a subset of the 30 seen classes. We start from $N\,{=}\,10$ and go on to $N\,{=}\,30$ seen classes by selecting, each time, the first $N$ classes in alphabetical order. For each $N$, we train all the networks used in our pipeline from scratch.
We use the W2V class prototypes.
For each $N$, we repeat the experiment 5 times and report the average scores obtained over these experiments.

We present in Figure~\ref{supp:fig:alphabetic} the global accuracy as a function of the number of seen classes. The list of selected seen classes is presented on the x-axis of the graph. 

We notice that the results are unstable for a number of seen classes below $16$ (\class{lamp}, \class{laptop}) or below $18$ (\class{mantel} and \class{person}). It is likely that the feature backbone is poorly trained, or that the generator is unable to ``align'' the space of class prototypes to the space of object representations when only a small number of seen classes is available. The performance stabilizes above $18$ seen classes, with a noticeable jump in performance around $29$ or $30$ classes. A possible explanation for this jump could be that the class \class{wardrobe} helps the alignment of the space of class prototypes onto the space of object representations because of some semantic closeness to the unseen classes \class{dresser} and \class{nightstand} and because of geometric similarities between point clouds belonging to these classes.
To validate this kind of hypotheses, more similar experiments would be required, changing the last, 30-th class, or changing the position at which the class \class{wardrobe} is included.

\subsection{ZSL confusion matrices}
\label{sec:confusion_zsl_ModelNet40}

As shown in Figure~\ref{supp:fig:cls_cm}(a) for the ZSL setting, a lot of different classes are wrongly predicted as \class{sofa} or \class{table}. It indicates that object representations generated for these two classes are similar to the representations of the classes that are falsely predicted as those two classes. In fact, these two classes are the equivalent of hubs as described in~\cite{cheraghian2019mitigating}, except that we use here representations in the object domain.

The confusion matrix also shows that the class \class{desk} is often mistaken for class \class{table}. The confusion in this case is probably caused by the fact that
the two classes are close both semantically at text level and geometrically/visually at 3D appearance level.  This also applies to class \class{nightstand}, that is often mistaken for class \class{dresser}.
A close textual semantics (embeddings) for these two classes could lead to generate geometrical features that are similar, which would make it hard for the classifier to tell apart instances of the two classes. The fact is that classes \class{nightstand} and \class{dresser} look alike in the ModelNet40 dataset as the CAD models are not in a metric scale.

We can note as well that nearly no object is ever predicted as \class{monitor}, even actual monitors themselves.
This could indicate that the generated object representation for the textual semantic embedding of this class is grossly wrong and does not come even close to the object representation of any other class.  A reason for this behavior could be that the semantic embedding of \class{monitor} is an outlier in the class prototype space. Consequently, no knowledge from scene classes can help discovering this unseen class.

\subsection{GZSL confusion matrices}
\label{ssec:gzsl_confusion_matrices}

In Figure~\ref{supp:fig:cls_cm}(b) and Figure~\ref{supp:fig:cls_cm}(c), we present the confusion matrices in the GZSL setting with $\beta=50$ and either $\epsilon=0$ or $\epsilon=0.995$.

By comparing the two confusion matrices, we can observe the influence of the calibrated stacking in the resulting distribution.
With $\epsilon=0$, predictions are good for the seen classes, but unseen classes are rarely predicted.
Using $\epsilon=0.995$, which is the parameter maximizing the harmonic mean on the validation set, we clearly see a shift of the predictions toward the unseen classes, that in fact greatly improves accuracy. Seen classes are however negatively impacted with this shift as their prediction becomes less sharp;
some of the seen classes that were correctly predicted with $\epsilon=0$ may then be predicted as one of the unseen classes.

We can see the parameter $\epsilon$ as a way to counter the natural tendency of the network to classify all objects as belonging to a seen (supervised) class, and re-balance the predictions between seen and unseen classes.

\begin{table*}
\centering
\begingroup
\renewcommand{\arraystretch}{0.9}
\setlength{\tabcolsep}{1.8pt}
\begin{tabular}
{c@{~~~}l c|c|c|c|c|ccc|ccc|ccc}
\toprule
\multirow{4}{*}{\rotatebox{90}{\!\!\!\!\!Dataset}}
& & &\multicolumn{3}{c|}{ZSL} & \multicolumn{10}{c}{GZSL}
\\
\cmidrule{4-16}
& & Gen- & \multirow{2}{*}{W2V} & \multirow{2}{*}{GloVe} & Glove&  Bias &\multicolumn{3}{c|}{W2V} & \multicolumn{3}{c|}{GloVe} & \multicolumn{3}{c}{GloVe + W2V}
\\ 
& Method & era- &  & & + W2V & reduc- & Acc. & Acc. & \multirow{2}{*}{HM} & Acc. &Acc. & \multirow{2}{*}{HM} & Acc.& Acc. & \multirow{2}{*}{HM}
\\
& & tive  & Acc. & Acc. & Acc.  & tion & $\seen$ & $\unseen$ &  & $\seen$ & $\unseen$ & & $\seen$ & $\unseen$ &
\\ 
\midrule
\multirow{6}{*}{\rotatebox{90}{\!\!\!ModelNet40}}
& f-CLSWGAN* \cite{xian2018feature} & \checkmark 
&20.7 & - & - &  & 76.3 & 3.7 & 7.0 & - & - & - & - & - & -  
\\
& CADA-VAE*  \cite{schonfeld2019generalized}  & \checkmark   
&23.0 & - & - &  & \bf 84.7 & 1.3 & 2.6 & - & - & - & - & - & -  
\\
& ZSLPC$^\dagger$ \cite{cheraghian2019zero} &  
& 28.0 & 20.9 & 20.5 &  & 40.1 & 22.5 & 28.8 & 49.2 & 18.2 & 26.6 & - & - & - 
\\
& MHPC \cite{cheraghian2019mitigating} &  
& \textbf{33.9} & 28.7 & -  & \checkmark  & 53.8 & 26.2 & 35.2 & \bf 53.8 & 25.7 & \textbf{34.8} & - & - & -
\\
& TZSLPC$^\ddagger$ \cite{cheraghian2020transductive} & 
& 23.5 & - & - & & 83.7 & 0.4 & 0.8 & - & - & - & - & - & - 
\\
& 3DGenZ (ours) & \checkmark
& 28.6 & \textbf{29.3} & \textbf{36.8} & \checkmark  & 48.8	& \textbf{29.3} & \textbf{36.6} & 44.7 & \textbf{28.4} & 34.7 & 47.8 & 36.5 & 41.3 
\\
\midrule
\multirow{6}{*}{\rotatebox{90}{\!\!McGill}}
& f-CLSWGAN* \cite{xian2018feature} & \checkmark  &10.2 & - & - &  & 75.3 & 2.3 & 4.5 & - & - & - & - & - & -  
\\
& CADA-VAE*  \cite{schonfeld2019generalized}  & \checkmark   &10.7 & - & - &  & \textbf{83.3} & 1.6 & 3.1 & - & - & - & - & - & -  
\\
& ZSLPC$^\dagger$ \cite{cheraghian2019zero} &   & 10.7 & 10.7 & \textbf{16.1} &  & - & - & - & - & - & - & - & - & - 
\\
& MHPC \cite{cheraghian2019mitigating} &  & 12.5 & \textbf{11.1} & -  & \checkmark  & - & - & - & - & - & - & - & - & -
\\
& TZSLPC$^\ddagger$ \cite{cheraghian2020transductive} &  & \textbf{13.0} & - & -  &   & 80.0 & 0.9 & 1.8 & - & - & - & - & - & -
\\
& 3DGenZ (ours) & \checkmark & 8.4 & 7.2 & 9.4 & \checkmark  & 50.5	& \textbf{7.2} & \textbf{12.5} & 48.9 & 6.4 & 11.3 & 49.6 & 8.6 & 14.5
\\
\midrule
\multirow{6}{*}{\rotatebox{90}{\!SHREC2015}}
& f-CLSWGAN* \cite{xian2018feature} & \checkmark  &5.2 & - & - &  & 74.2 & 0.8 & 1.6 & - & - & - & - & - & -
\\
& CADA-VAE*  \cite{schonfeld2019generalized}  & \checkmark   &\textbf{6.2} & - & - & - & 80.0 & 1.7 & 3.3 & - & - & - & - & - & -
\\
& ZSLPC$^\dagger$ \cite{cheraghian2019zero} &   & 5.2 & 3.6 & \textbf{6.8} &  & - & - & - & - & - & - & - & - & - 
\\
& MHPC \cite{cheraghian2019mitigating} &  & \textbf{6.2} & \textbf{4.1} & -  & \checkmark  & - & - & - & - & - & - & - & - & -
\\
& TZSLPC$^\ddagger$ \cite{cheraghian2020transductive} &  & 5.2 & - & -  &   & \textbf{82.1} & 0.9 & 1.8 & - & - & - & - & - & -
\\
& 3DGenZ (ours) & \checkmark & 4.9 & \textbf{4.1} & 4.9& \checkmark  & 54.1	& \textbf{4.3} & \textbf{8.0} & 47.2 & 3.9 & 7.2 & 50.9 & 4.6 & 8.4
\\
\bottomrule
\multicolumn{1}{l}{}\\[-4pt]
\end{tabular}
\endgroup
\caption{ZSL and GZSL classification results (in \%) on ModelNet40 \cite{wu20153d}, McGill~\cite{siddiqi2008retrieving} and SHREC2015~\cite{10.2312:3dor.20151064}. For a fair comparison, we report results based on the same PointNet backbone. Results are averaged over 20 runs for 3DGenZ (not other methods). Missing figures are due to code unavailable and previous publications not evaluating with all kinds of word embeddings. \newline\mbox{}\qquad *: adaptation of 2D methods to 3D point clouds, implemented in~\cite{cheraghian2020transductive}. \newline\mbox{}\qquad $^\dagger$: best reported variant in \cite{cheraghian2019zero}, i.e., PointNet + NetVlad. \newline\mbox{}\qquad $^\ddagger$: inductive baseline reported in \cite{cheraghian2020transductive}.}
\label{supp:tab:zsl_all_gzsl_overview}
\end{table*}

\subsection{Classification results on more datasets}

Section~4.4.1 and Table 1 of the main paper compare our method to state-of-the-art ZSL and GZSL classification on the classic ModelNet40 \cite{wu20153d} dataset. We complete here the comparison by experimenting as well on McGill~\cite{siddiqi2008retrieving} and SHREC2015~\cite{10.2312:3dor.20151064} datasets.

Following \cite{cheraghian2019zero, cheraghian2019mitigating, cheraghian2020transductive}, we consider ModelNet40 classes as seen classes, and use unseen classes from McGill (14 classes, 115 examples) and SHREC2015 (30 classes, 192 examples).
To be comparable with the literature, we follow \cite{cheraghian2019mitigating} when adapting theses datasets for the ZSL and GZSL tasks. It means in particular that classes appearing in the seen classes of ModelNet40 are removed from the McGill and SHREC2015 dataset, and are not used as unseen classes. 
Besides, to be comparable with the experiments in \cite{cheraghian2019zero, cheraghian2019mitigating, cheraghian2020transductive}, we use the same PointNet backbone as we already used for the experiments with the ModelNet40 unseen classes. Furthermore, we use the same hyperparameters for the bias reduction as we used for the ModelNet40 unseen classes, as the same seen classes are used and no additional cross-validation is necessary.

Results are shown in Table~\ref{supp:tab:zsl_all_gzsl_overview}. (For a complete overview, we also recall in Table~\ref{supp:tab:zsl_all_gzsl_overview} the results on ModelNet40, that were already provided in Table~1 of the main paper.) Please note that missing figures are due to code unavailable and to previous publications not evaluating with all kinds of word embeddings.

These test datasets are  difficult challenges for ZSL and GZSL as the 30 classes of SHREC2015 as well as 10 of the 14 classes in the McGill datasets are animals, whereas the ModelNet40 seen classes used for training do not contain a single animal and focus on man-made objects. This is probably a reason why, in Figure~3 of~\cite{cheraghian2019zero}, the t-SNE visualisation of the McGill and ModelNet40 word representations looks quite disjoint. Morevoer, it can be noted that the number of test examples for SHREC2015 and McGill is quite low compared to the 908 test examples for unseen classes in ModelNet40. More variation from one method to another can thus be expected with these two datasets.

Indeed, compared to the results with ModelNet40 unseen classes, an overall drop in Top-1 Acc.\ and HM can be observed on McGill and SHREC2015 unseen classes, both in our results and with the other methods. It corroborates the difficulty of the task mentioned above.

For the ZSL task, we are a bit below the state of the art with word2vec (W2V) embeddings, but we are comparable with Glove embeddings (better on ModelNet40, not as good on McGill, equal on SHREC2015). 
For the GZSL task, which is also relevant for semantic segmentation, we see that our framework with bias reduction achieves state-of-the-art results (HM) on all three datasets with W2V embeddings, and is comparable to the best method using GloVe embeddings (on ModelNet40, which is the only dataset tested in the literature with these embeddings).

(In a recent unpublished work \cite{cheraghian2021zero}, better results are obtained on McGill with a different backbone network. However, as the choice of the backbone network, for the same ZSL or GZSL method, has a large impact on performance, as can be seen in Tables 2 and~3 of~\cite{cheraghian2021zero}, these results are not directly comparable to those reported in Table~\ref{supp:tab:zsl_all_gzsl_overview} nor in the main paper.)

These ZSL and GZSL classification results validate our approach and suggest it is relevant to use it as a general framework to also derive a semantic segmentation method.

\begin{table*}
\centering
\begingroup
\renewcommand{\arraystretch}{0.9}
\setlength{\tabcolsep}{1.8pt}
\begin{tabular}
{l|c|ccc|ccc|ccc}
\toprule
& \multicolumn{10}{c}{GZSL} \\
\cmidrule{2-11}
\multicolumn{1}{c|}{Method} & Bias&\multicolumn{3}{c|}{W2V} & \multicolumn{3}{c|}{GloVe} & \multicolumn{3}{c}{GloVe + W2V}\\ 
  & reduc- & Acc. & Acc. & \multirow{2}{*}{HM} & Acc. &Acc. & \multirow{2}{*}{HM} & Acc.& Acc. & \multirow{2}{*}{HM}\\
 & tion & $\seen$ & $\unseen$ &  & $\seen$ & $\unseen$ & & $\seen$ & $\unseen$ & \\ 
\midrule
3DGenZ w/o bias reduct.  &   & \textbf{79.3} & 9.96 & 17.6 & \textbf{79.7} & 12.1 & 21.0 & \textbf{79.0} & 13.4 & 22.8\\
3DGenZ w/ our bias reduct. & \checkmark  & 48.8	& \textbf{29.3} & \textbf{36.6} & 44.7 & \textbf{28.4} & \textbf{34.7} & 47.8 & \textbf{36.5} & \textbf{41.3} \\
\bottomrule
\multicolumn{1}{l}{}\\[-7pt]
\end{tabular}
\endgroup
\caption{Ablation study: GZSL classification with 3DGenZ on ModelNet40 with and without our bias reduction mechanism.}
\label{tab:zsl_gzsl_overview_bias_wobias}
\vspace{2mm}
\end{table*}

\begin{table*}[t]
\centering
\begin{tabular}{l|l|l|l|l|l}
\toprule
Datasets & Split 1                                                                          & Split 2                                                                    & Split 3                                                                               & Split 4                                                 & Split 5                                                 \\ \hline
S3DIS         & \begin{tabular}[c]{@{}l@{}}Door, \\ Bookcase\end{tabular}                        & \begin{tabular}[c]{@{}l@{}}Bookcase, \\ Board\end{tabular}                 & \begin{tabular}[c]{@{}l@{}}Door,  \\ Table\end{tabular}                               & \begin{tabular}[c]{@{}l@{}}Table, \\ Chair\end{tabular} & \begin{tabular}[c]{@{}l@{}}Chair, \\ Board\end{tabular} \\ \hline
ScanNet         & \begin{tabular}[c]{@{}l@{}}Chair, \\ Table, \\ Cabinet\end{tabular}                        & \begin{tabular}[c]{@{}l@{}}Counter, \\ Bathtub, \\ Sink\end{tabular}                 & \begin{tabular}[c]{@{}l@{}}Counter,  \\ Table, \\ Sink\end{tabular}                               & \begin{tabular}[c]{@{}l@{}}Chair, \\ Cabinet, \\ Bathtub\end{tabular} & - \\ \hline
SemanticKITTI & \begin{tabular}[c]{@{}l@{}}Other-vehicle,\\ Person, \\ Motorcyclist\end{tabular} & \begin{tabular}[c]{@{}l@{}}Bicycle, \\ Person,\\ Other-ground\end{tabular} & \begin{tabular}[c]{@{}l@{}}Other-vehicle,\\ Motorcyclist,\\ Other-ground\end{tabular} & -                                                       & -                                                      \\
\bottomrule
\multicolumn{1}{l}{}\\[-7pt]
\end{tabular}
\caption{Classes used for validation in the different cross-validation splits for the semantic segmentation task.}
\label{supp:tab:split_semantic_segmentation_tables}
\end{table*}

\subsection{Ablation study of bias reduction}

Section~4.3 and Figure~3 of the main paper provide an ablation and a parameter range study for ZSL classification (on ModelNet40), and for GZSL semantic segmentation (on S3DIS, ScanNet and SemanticKITTI). We complete here this ablation study with the case of GZSL classification.

Table \ref{tab:zsl_gzsl_overview_bias_wobias} reports the results for GZSL classification on ModelNet40 for our 3DGenZ framework with and without our bias reduction method. It can be seen clearly that our bias reduction improves the HM and the accuracy of unseen classes for all kinds of word embeddings. Because bias reduction provides a trade-off, this improvement comes with a drop of the performance in the seen classes, as already described in Section~\ref{ssec:gzsl_confusion_matrices}.

\section{Semantic segmentation}
\label{sup:sec:semantic_segmentation}

\subsection{Cross-validation splits}
\label{sec:cross_val_seg}

Cross-validation is done with 5 splits on S3DIS, 4 splits on ScanNet and 3 splits on SemanticKITTI.
Following \cite{Bucher_2017_ICCV}, we select 20\% of the seen classes in each split as validation classes with a minimum of at least 2 classes.
Therefore, we have 2, 3 and 3 selected validation classes in S3DIS, ScanNet and SemanticKITTI, respectively.

For each validation split, as we are in the inductive zero-shot setting, the feature backbone is trained
using only seen-class data that do not include classes selected as validation classes, and the validation evaluations are done only on the validation classes. We would also like to highlight again that the unseen classes for testing are \emph{not} used in the validation process.

As explained in the main part of the paper, frequently appearing classes in the semantic segmentation datasets cannot be used as validation classes because removing them would drastically reduce the size of the training dataset. We present in Table~\ref{supp:tab:split_semantic_segmentation_tables} our choices of validation classes. These chosen validation classes avoid reducing too much the amount of remaining training data.

\begin{table*}[t]
\centering
\newcommand*\rotext{\multicolumn{1}{R{45}{1em}}}
\setlength{\tabcolsep}{4pt}
\begin{tabular}{lc|rrrrrrrrr|>{\columncolor{LightGrey}}r>{\columncolor{LightGrey}}r>{\columncolor{LightGrey}}r>{\columncolor{LightGrey}}r|r}
\toprule
& & \multicolumn{9}{c|}{seen classes} & \multicolumn{4}{>{\columncolor{LightGrey}}c|}{\textbf{unseen classes}} & \cellcolor{HL}\\
\multicolumn{2}{c|}{\stackbox{S3DIS\\split\\\mbox{}}}
& \rotext{Board} & \rotext{Bookcase} & \rotext{Ceiling} & \rotext{Chair} & \rotext{Clutter} & \rotext{Door} & \rotext{Floor} & \rotext{Table} & \rotext{Wall} & \rotext{\textbf{Beam}} & \rotext{\textbf{Column}} & \rotext{\textbf{Sofa}} & \rotext{\textbf{Window}} & \stackbox{Hm\\IoU} \\ \hline
FSL & IoU & 
53.9 & 54.4 & 96.5 & 75.9 & 66.0 & 78.7 & 96.0 & 70.3  & 74.1 & 63.1 & 10.2 & 54.1 & 72.4 & \cellcolor{HL}59.6 \\ 
3DGenZ & IoU & 
19.1 & 34.1 & 92.8 & 56.3 & 39.2 & 25.4 & 91.5 & 57.3  & 62.3 & 13.9 & ~~2.4 & 4.9 & 8.1 & \cellcolor{HL}12.9 \\ \hline
3DGenZ & Acc. & 
19.7 & 39.8 & 96.9 & 58.7 & 43.5 & 25.8 & 92.9 & 61.9 & 80.3 & 20.0 & 9.1 & 62.4 & 23.7 & \cellcolor{HL}- \\
\bottomrule
\multicolumn{1}{l}{}\\[-7pt]
\end{tabular}
\caption{Classwise GZSL semantic segmentation performance (\%) on the S3DIS split: fully-supervised learning (FSL), i.e., training using annotations for both seen and unseen classes, as upper bound, and GZSL with 3DGenZ with respect to unseen classes (in bold face).}
\label{sup:tab:s3dis_acc_miou}
\end{table*}

\subsection{Baselines for semantic segmentation}

To our knowledge, this work is the first\footnote{An unpublished report on this topic was recently made public \cite{liu2021segmenting}. However, it operates in a different setting as the unseen classes are present at training time, although unlabeled. In our inductive setting, strictly no unseen class can seen at training time, which makes the task substantially more difficult and which also reduces the size of training data to satisfy this constraint.} to address zero-shot semantic segmentation for point clouds.
To show the efficiency of our proposed approach, we designed two baselines based on previous work for zero-shot classification.

As baselines, we adapted the ZSLPC \cite{cheraghian2019zero} and DeViSE \cite{bucher2019zero} methods to semantic segmentation. However, a direct adaption did not produce valuable results as we observed a strong prediction bias towards the seen classes on all datasets we experimented with (see lines for ZSLPC-Seg* and DeViSe-3DSeg* in Table 2 of the paper). A bias reduction mechanism was thus also needed for these two baseline methods.

However, these methods use a different paradigm than ours: they base classification on a nearest-neighbor search in the space of class prototypes, whereas we produce classification scores and pseudo-probabilities (after softmax) via a trained classifier. As a result, it was not possible to apply the same bias reduction techniques that we used (class-dependent weighting and calibrated stacking). Nevertheless, we tried to rebalance unseen classes by reducing the distance to prototypes of unseen classes by a constant value; it is somehow similar to the calibrated stacking we are using, where the pseudo-probability of unseen classes is increased by~$\epsilon$, but it is in a totally different and much larger space. Unfortunately, it did not lead to valuable results.

To construct meaningful baselines, we thus had to design a more complex bias reduction technique.
To reduce the bias towards seen classes, we proceed as follows: we search the $K$-nearest-neighbors in the space of class prototypes; if a prototype of an unseen class is present among these $K$ neighbors, we pick the class of the nearest prototype of unseen class; otherwise, we pick the class of the nearest prototype. We call the corresponding methods ZSLPC-Seg and DeViSe-3DSeg, respectively.

\begin{table}[t]
\begin{center}\setlength{\tabcolsep}{4pt}
\begin{tabular}{l|ccc}
\toprule
Method \textbackslash\ Dataset & S3DIS & ScanNet & SemanticKITTI \\
\midrule
ZSLPC-Seg       & 5 & 2 & 5 \\
DeViSe-3DSeg    & 7 & 2 & 5 \\
\bottomrule
\end{tabular}
\end{center}
\vspace*{-1.5mm}
\caption{Best value of bias reduction parameter $K$ for the baseline GZSL semantic segmentation methods.}
\label{supp:tab:K}
\end{table}

As described in the paper, we select the best performing value for $K$ depending on the dataset (see Table~\ref{supp:tab:K}).

\subsection{Classwise performance on segmentation}

\begin{table*}[t]
\centering
\newcommand*\rotext{\multicolumn{1}{R{45}{1em}}}
\setlength{\tabcolsep}{1.7pt}
\begin{tabular}{@{}l@{~}c|rrrrrrrrrrrrrrrr|>{\columncolor{LightGrey}}r>{\columncolor{LightGrey}}r>{\columncolor{LightGrey}}r>{\columncolor{LightGrey}}r|r@{}}
\toprule
&& \multicolumn{16}{c|}{{seen classes}} & \multicolumn{4}{>{\columncolor{LightGrey}}c|}{\textbf{unseen classes}} & \cellcolor{HL} \\
\multicolumn{2}{c|}{\stackbox{ScanNet\\split\\\mbox{}}}
& \rotext{Bathtub} & \rotext{Bed} & \rotext{Cabinet} & \rotext{Chair} & \rotext{Counter} & \rotext{Curtain} & \rotext{Door} & \rotext{Floor}  & \rotext{Other furni\rlap{ture}} & \rotext{Picture} & \rotext{Refrigera\rlap{tor}} & \rotext{Shower cur\rlap{tain}} & \rotext{Sink} & \rotext{Table} & \rotext{Wall} & \rotext{Window} & \rotext{\textbf{Bookshelf}} & \rotext{\textbf{Desk}} & \rotext{\textbf{Sofa}} & \rotext{\textbf{Toilet}} & \stackbox{Hm\\IoU} \\
\hline
FSL & IoU & 
58.0	&67.5&	21.2&	75.5&	12.0&	35.2&	13.6&	96.5&	20.6&	10.7&	39.9&	63.3&	34.2&	59.5&	81.1&	4.8&	56.9&	30.0&	57.4&	63.4 & \cellcolor{HL}47.2
\\ 
3DGenZ & IoU & 
64.9&	44.0&	16.9&	63.2&	15.3&	33.8&	10.4&	91.0&	10.1&	4.3&	26.1&	0.2&	27.5&	43.1&	71.3&    2.8&	6.3&	3.3&	13.1&	8.1 & \cellcolor{HL}12.5
\\ \hline
3DGenZ & Acc. & 
75.7&	68.3&	27.6&	78.1&	35.4&	40.2&	12.1&	97.4&	18.5&	5.1&	31.3&	0.3&	44.3&	56.0&	83.2&	3.1&	13.4&	5.9&	49.6&	26.3 & \cellcolor{HL}-
\\

\bottomrule
\multicolumn{1}{l}{}\\[-7pt]
\end{tabular}
\caption{Classwise GZSL semantic segmentation performance (\%) on the ScanNet split: fully-supervised learning (FSL), i.e., training using annotations for both seen and unseen classes, as upper bound, and GZSL with 3DGenZ with respect to unseen classes (in bold face).}
\label{sup:tab:sn_acc_miou}
\end{table*}

\subsubsection{Performance details on S3DIS}

\begin{figure}[t]
    \centering
    \includegraphics[width=0.85\columnwidth]{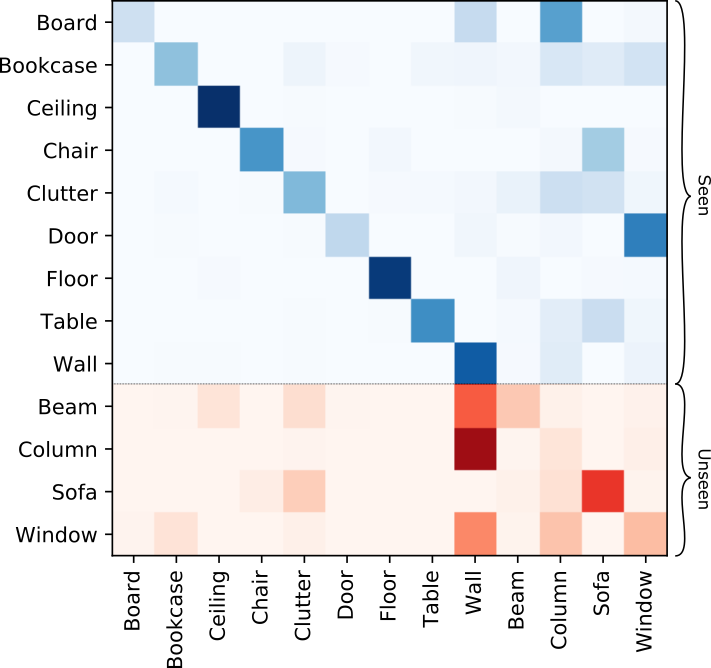}
    \vspace*{-0mm}
    \caption{Confusion matrix for the GZSL semantic segmentation of S3DIS with 3DGenZ. 
    The red color map is for unseen classes, and the blue one for seen classes.
    Each row shows the distribution of predictions of one class: the darker the color is, the more often points of this class are predicted as the class of the column.}
    \label{supp:fig:seg_cm_s3dis}
\end{figure}

The GZSL semantic segmentation method presented in this paper achieves a HmIoU of 12.9\% on the test data of S3DIS (Table 1 of the main paper). The detailed classwise results are shown here in Table~\ref{sup:tab:s3dis_acc_miou}. 

According to per-class IoU, the best performing unseen class is \class{Beam}, and the worst is \class{Column}. However, a relatively bad performance of the class \class{Column} can also be seen in the full-supervised learning scenario (FSL). This could indicate that this class is based on visual features that are hard to differentiate from other classes.

Regarding classwise accuracy, \class{Sofa} also performs very well. Yet, a large Acc with a low IoU is the indication that points that should be labeled with another class (than \class{Sofa}) are actually mispredicted as \class{Sofa}.
The per-row normalized confusion matrix in Figure~\ref{supp:fig:seg_cm_s3dis} supports this assumption, as $35\%$ of the \class{Chairs} are predicted as \class{Sofas}. A reason could be that while substantially different geometrically, these two classes are close regarding textual semantic, hence word embeddings.

It can also be seen in the confusion matrix that the classes \class{Beam}, \class{Column} and \class{Window} are often falsely predicted as \class{Wall}, which could be due to the fact that all these classes similarly feature large flat smooth surfaces.

The confusion matrix additionally shows that, probably due to remaining weight issues, some seen classes have a tendency to be predicted with the label of an unseen class, although for these seen classes the classifier is presented with the actual 3D features, as opposed to generated ones. For example, the class \class{Board} is falsely predicted as \class{Column} in 54\% of the cases. This effect probably also contributes to many \class{Chair} points being predicted as \class{Sofa}.

\begin{figure}[t]
    \centering
    \vspace*{-2mm}
    \includegraphics[width=\columnwidth]{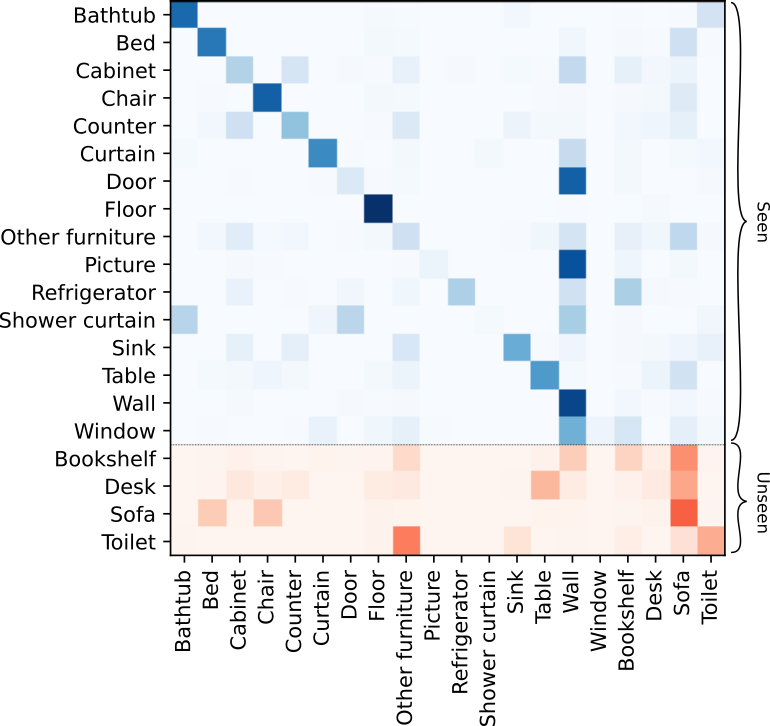}
    \vspace*{-3mm}
    \caption{Confusion matrix for the GZSL semantic segmentation of ScanNet with 3DGenZ. 
    The red color map is for unseen classes, and the blue one for seen classes.
    Each row shows the distribution of predictions of one class: the darker the color is, the more often points of this class are predicted as the class of the column.}
    \label{supp:fig:seg_cm_scannet}
    \vspace*{-2mm}
\end{figure}

\subsubsection{Performance details on ScanNet}

To complete the results presented in Table~1 of the paper, the classwise performance in the FSL and GZSL settings on the test set of ScanNet are reported in Table~\ref{sup:tab:sn_acc_miou}. The confusion matrix for the GZSL scenario is shown in Figure~\ref{supp:fig:seg_cm_scannet}.

Like for S3DIS, a number of seen classes are wrongly classified as \class{Wall}.  Again, it could be due to large flat surfaces that are shared by all of the misclassified classes.

In the unseen classes, an intriguing observation is the bias of the \class{Desk} and \class{Bookshelf} classes towards the \class{Sofa} class. We hypothesize that features extracted by the backbone for examples of \class{Bookshelf} and \class{Desk} are closer to generated representations for \class{Sofa} than to generated representations of their own class.
As several seen classes are also wrongly classified as \class{Sofa}, we suspect that the \class{Sofa} class accumulates hard-to-classify examples besides examples of its own class, in a hub-like effect~\cite{cheraghian2019mitigating} already observed for classification (see Section~\ref{sec:confusion_zsl_ModelNet40}).

The unseen class \class{Desk} is also often misclassified as the seen class \class{Table}. This ambiguity exists both on the geometrical and on the semantic level.

Another observation is that, compared to the SemanticKITTI and the ModelNet40 datasets, the bias correction is less strong towards unseen classes.

\begin{figure}[t]
    \centering
    \vspace*{-2mm}
    \includegraphics[width=\columnwidth]{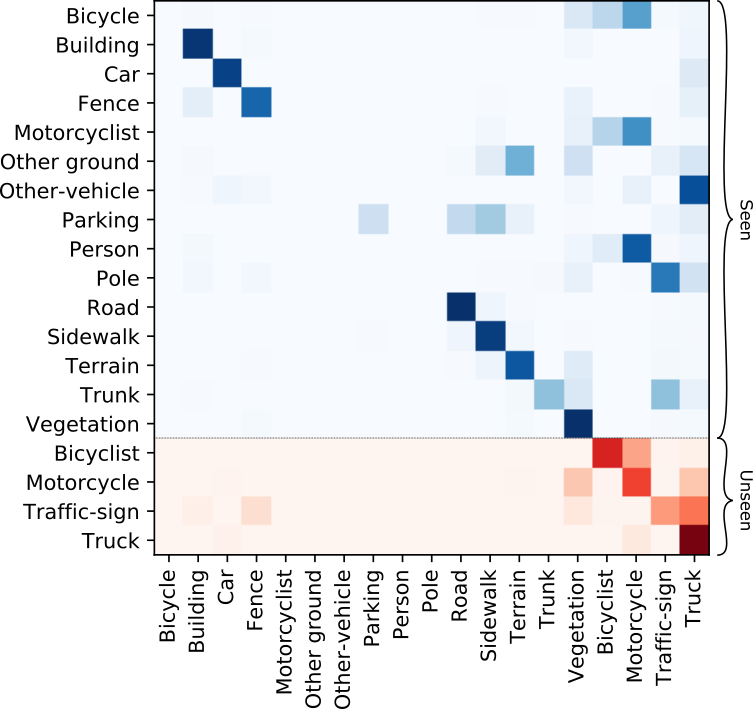}
    \vspace*{-3mm}
    \caption{Confusion matrix for the GZSL semantic segmentation of SemanticKITTI with 3DGenZ. 
    The red color map is for unseen classes, the blue one for seen classes.
    Each row shows the distribution of predictions of one class: the darker the color is, the more often points of this class are predicted as the class of the column.}
    \label{supp:fig:seg_cm_sk}
\end{figure}

\subsubsection{Performance details on SemanticKITTI}

\begin{table*}[t]
\centering
\newcommand*\rotext{\multicolumn{1}{R{45}{1em}}}
\setlength{\tabcolsep}{2.2pt}
\begin{tabular}{l@{~}c|rrrrrrrrrrrrrrr|>{\columncolor{LightGrey}}r>{\columncolor{LightGrey}}r>{\columncolor{LightGrey}}r>{\columncolor{LightGrey}}r|r}
\toprule
(a)&& \multicolumn{15}{c|}{{seen classes}} & \multicolumn{4}{>{\columncolor{LightGrey}}c|}{\textbf{unseen classes}}  & \cellcolor{HL}\\
\multicolumn{2}{c|}{\raisebox{2.5mm}{\stackbox{\smash{Semantic}\\KITTI\\ \small split\,1 (main)}}}
& \rotext{Bicycle} & \rotext{Building} & \rotext{Car} & \rotext{Fence} & \rotext{Motorcyclist} & \rotext{Other ground} & \rotext{Other vehicle} & \rotext{Parking}  & \rotext{Person} & \rotext{Pole} & \rotext{Road} & \rotext{Sidewalk} & \rotext{Terrain} & \rotext{Trunk} & \rotext{Vegetation} & \rotext{\textbf{Bicyclist}} & \rotext{\textbf{Motorcycle}} & \rotext{\textbf{Traffic sign}} & \rotext{\textbf{Truck}} & \stackbox{Hm\\IoU} \\
\hline
FSL & IoU & 
42.0 & 88.6 & 93.6 & 65.8 & 0.0 & 2.7 & 41.1 & 28.9 & 69.7 & 63.7 & 89.4 & 77.1 & 70.5 & 70.7 & 87.5 & 74.4 & 58.6 & 26.7 & 41.6 & \cellcolor{HL}54.5 \\ 
3DGenZ & IoU & 
0.0 & 87.3 & 86.9 & 61.8 & 0.0 & 0.0 & 0.0 & 18.6 & 0.0 & 0.0 & 88.8 & 78.6 & 73.6 & 38.2 & 87.8 & 28.0 & 11.5 & 0.9 & 2.6 & \cellcolor{HL}17.1 \\
\hline
3DGenZ & Acc.  & 
0.0 & 91.5 & 87.4 & 74.8 & 0.0 & 0.0 & 0.0 & 19.9 & 0.0 & 0.0 & 93.3 & 89.2 & 79.8 & 38.6 & 94.0 & 66.8 & 57.2 & 32.7 & 90.8 & \cellcolor{HL}-\\
\bottomrule
\multicolumn{1}{l}{}\\[-7pt]
\end{tabular}
\setlength{\tabcolsep}{2.2pt}
\begin{tabular}{l@{~}c|rrrrrrrrrrrrrrr|>{\columncolor{LightGrey}}r>{\columncolor{LightGrey}}r>{\columncolor{LightGrey}}r>{\columncolor{LightGrey}}r|r}
\toprule
(b)&& \multicolumn{15}{c|}{{seen classes}} & \multicolumn{4}{>{\columncolor{LightGrey}}c|}{\textbf{unseen classes}} & \cellcolor{HL}  \\
\multicolumn{2}{c|}{\raisebox{2.5mm}{\stackbox{\smash{Semantic}\\KITTI\\ \small split\,2 (altern.)}}}
& \rotext{Bicyclist
} & \rotext{Building} & \rotext{Car} & \rotext{Fence} & \rotext{Motorcycle
} & \rotext{Other ground} & \rotext{Other vehicle} & \rotext{Parking}  & \rotext{Person} & \rotext{Pole} & \rotext{Road} & \rotext{Sidewalk} & \rotext{Terrain} & \rotext{Trunk} & \rotext{Vegetation} & \rotext{\textbf{Bicycle
}} & \rotext{\textbf{Motorcyclist
}} & \rotext{\textbf{Traffic sign}} & \rotext{\textbf{Truck}} & \stackbox{Hm\\IoU} \\
\hline
FSL & IoU & 
74.4 & 88.6 & 93.6 & 65.8 & 58.6 & 2.7 & 41.1 & 28.9 & 69.7 & 63.7 & 89.4 & 77.1 & 70.5 & 70.7 & 87.5 & 42.0 & 0.0 & 26.7 & 41.6 & \cellcolor{HL}38.8 \\ 
3DGenZ & IoU & 0.0&    84.5&   78.9&   53.5&   3.9&    0.0&    0.0&    21.8&   0.0&    0.0&    85.4&   72.6&    67.8&   50.1&   87.9&   0.0&    0.3&    3.0&    2.0 & \cellcolor{HL}12.7\\
\hline
3DGenZ & Acc.  & 0.0&    91.6&   80.1&   73.3&   14.9&   0.0&    0.0&   0.0&    0.0&    91.7&   87.3&   25.6&   75.3&   55.1&   95.4&   0.0&    51.1&   25.6&   30.8 & \cellcolor{HL}-\\
\bottomrule
\multicolumn{1}{l}{}\\[-7pt]
\end{tabular}
\setlength{\tabcolsep}{2.2pt}
\begin{tabular}{l@{~}c|rrrrrrrrrrrrrrr|>{\columncolor{LightGrey}}r>{\columncolor{LightGrey}}r>{\columncolor{LightGrey}}r>{\columncolor{LightGrey}}r|r}
\toprule
(c)&& \multicolumn{15}{c|}{{seen classes}} & \multicolumn{4}{>{\columncolor{LightGrey}}c|}{\textbf{unseen classes}} & \cellcolor{HL} \\
\multicolumn{2}{c|}{\raisebox{2.5mm}{\stackbox{\smash{Semantic}\\KITTI\\ \small split\,3 (altern.)}}}
& \rotext{Building} & \rotext{Car} & \rotext{Fence} & \rotext{Other ground} & \rotext{Other vehicle} & \rotext{Parking}  & \rotext{Person} & \rotext{Pole} & \rotext{Road} & \rotext{Sidewalk} & \rotext{Terrain} & \rotext{Traffic sign}& \rotext{Truck} &\rotext{Trunk} & \rotext{Vegetation} &
\rotext{\textbf{Motorcycle}} &
\rotext{\textbf{Motorcyclist}}&
\rotext{\textbf{Bicyclist}}& 
\rotext{\textbf{Bicycle}} & \stackbox{Hm\\IoU} \\
\hline
FSL & IoU & 
88.6 & 93.6 & 65.8 & 2.7 & 41.1 & 28.9 & 69.7 & 63.7 & 89.4 & 77.1 & 70.5 & 26.7 & 41.6 & 70.7 & 87.5 & 58.6 & 0.0 & 74.4& 42.0 & \cellcolor{HL}51.0  \\ 
3DGenZ & IoU & 82.4&   82.8&   47.2&   0.0&    0.0&    15.3&   0.0&    0.0&    82.9&   70.2&   0.0&    66.9&   0.0&    0.1&    88.5&    0.9&    1.9&    0.1&   0.0 &\cellcolor{HL}1.4\\
\hline
3DGenZ & Acc.  & 94.4&   82.9&   57.7&   0.0&    0.0&   17.6&   0.0&    0.0&    92.1&   82.3&   0.0&   77.1&   0.0&    0.1&    96.0&    18.3&   74.5&    3.0&   0.0 & \cellcolor{HL}-\\
\bottomrule
\multicolumn{1}{l}{}\\[-7pt]
\end{tabular}
\caption{Classwise semantic segmentation performance (\%) on SemanticKITTI using main split 1 (a) or alternative splits~2 (b) and~3~(c): fully-supervised learning (FSL), i.e., training using annotations for both seen and unseen classes, as upper bound performance, and GZSL with 3DGenZ w.r.t.\ unseen classes.}
\label{sup:tab:sk_acc_miou_all_splits}
\vspace*{-1.3mm}
\end{table*}

Table~\ref{sup:tab:sk_acc_miou_all_splits}(a) provides the classwise semantic segmentation performance for our method on the SemanticKITTI dataset (main split). As for S3DIS and ScanNet, the Acc is much larger than the IoU for unseen classes. The confusion matrix in Figure~\ref{supp:fig:seg_cm_sk} confirms here as well that it is due to points of seen classes being predicted as some unseen class.
For example, the instances of the seen class \class{Other vehicle} are predicted in 91\% of the cases as the unseen class \class{Truck}, and the seen class \class{Person} is predicted in 78\% of the cases as the unseen class \class{Motorcycle}. 

There is also a cluster of classes whose textual semantics and 3D appearance are strongly connected, which might cause some confusion.
These are the classes \class{Bicycle}, \class{Motorcycle}, \class{Bicyclist}, \class{Motorcyclist} and \class{Person}. Classes \class{Motorcyclist} and \class{Bicyclist} are used for the person as well as the motorcycle if this person is on a (motor)bike. The prediction on the seen classes \class{Bicycle}, \class{Person} and \class{Motorcyclist} is in the majority of the cases distributed between the unseen classes of \class{Motorcycle} and \class{Bicyclist}.
For the mentioned seen classes, it is very harmful; in fact, they have an IoU of 0\%. 

It illustrates that, in this kind of setting, the class-dependent weighting and the calibrated stacking may turn the bias towards seen classes into a bias towards unseen classes, although their parameters $\beta$ and $\epsilon$ are specifically and systematically adapted to the dataset. However, these bias-reduction parameters are set by cross-validation using validation-unseen classes that are unrelated test-unseen classes, which could be an issue.

\begin{figure*}[t]
\centering\setlength{\tabcolsep}{4pt}
\vspace{-2mm}
\begin{tabular}[t]{@{}c|c|c@{}}
\raisebox{-.5\height}{\includegraphics[width=0.32\linewidth,trim=110 30 0 20,clip]{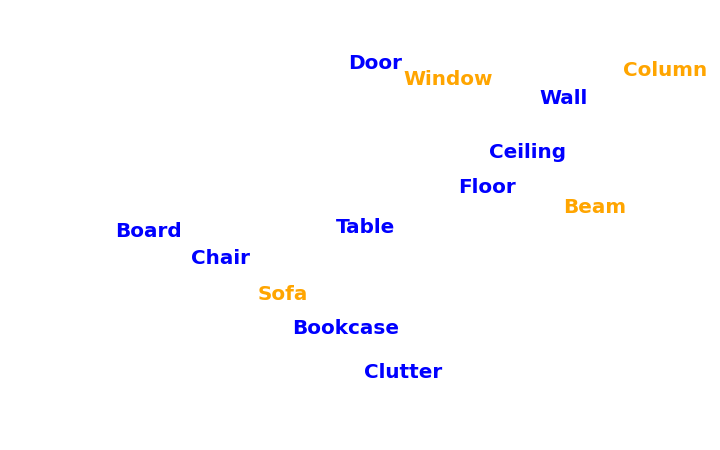}}&
\raisebox{-.5\height}{\includegraphics[width=0.32\linewidth,trim=110 30 0 20,clip]{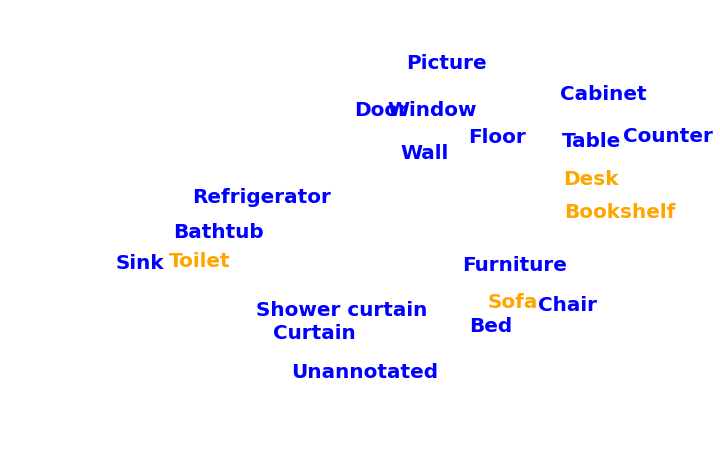}}&
\raisebox{-.5\height}{\includegraphics[width=0.32\linewidth,trim=110 30 0 20,clip]{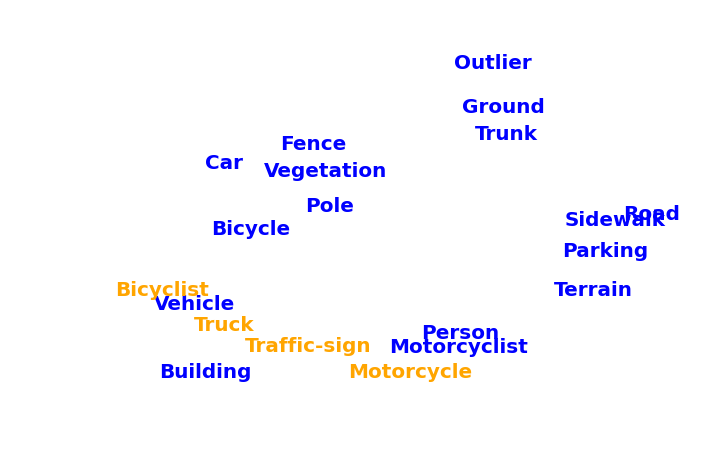}}
\\
\small (a) S3DIS & \small (b) ScanNet & \small (c) SemanticKITTI (split 1 for color)
\end{tabular}
\vspace*{1mm}
\caption{t-SNE \cite{van2014accelerating}  visualizations of the W2V+GloVe class prototypes for semantic segmentation datasets (\textcolor{blue}{\bf blue seen}, \textcolor{orange}{\bf orange unseen}).}
\label{supp:fig:ss_tsne_representation}
\end{figure*}

\subsection{Alternative splits}

As the 3D ZSL segmentation task is new, no benchmark is available to evaluate our method. We thus had to make our own benchmarks, creating class splits in existing 3D semantic segmentation datasets and curating data for inductive ZSL (strictly no unseen class in training data).

To create these splits, as already stated, one of the concerns is to keep as much training data as possible, which favors less represented classes as unseen classes. However, the choice of unseen classes also defines the difficulty of the benchmarks. Section~4.1 of the paper presents the rationale of the split choices for S3DIS, ScanNet and SemanticKITTI. As we hope that our benchmarks will be used for further research, we define here two additional splits on SemanticKITTI, that we see as even more challenging than the split we present in the paper, and we evaluate our method on them.

In the alternative split 2 for SemanticKITTI, we start from the split presented in the paper (main split~1) and we replace the role of bicycles and motorbikes, i.e., \class{motorbikes} and \class{bicyclists} are now seen, while \class{motorcyclists} and \class{bicycles} are now unseen. The motivation for this split is exactly the same as the one we describe in the paper, i.e., allowing to leverage on closely related classes. 
However, we see this other split as more challenging, at least for our backbone, as the IoU for \class{bicycle} is comparatively lower in the FSL scenario; furthermore, the IoU is even 0.0\% for \class{motorcyclist}. Results for this alternative split are reported in Table~\ref{sup:tab:sk_acc_miou_all_splits}(b). Our model achieves on this split an HmIoU that is 4.4 points lower than the HmIoU of 17.1\% achieved on the split used in the paper (split~1). We assume it is linked to the intrinsic difficulty of classifying \class{bicycle} and \class{motorcyclist}, as highlighted in the FSL scenario.

An even more difficult scenario is the selection of classes \class{Bicycle}, `\class{Bicyclist}, \class{Motorcycle} and \class{Motorcyclist} as unseen, given that these four classes are all semantically and geometrically very close and that it is difficult to tell them apart. This assumption is confirmed by the results on this alternative split~3, reported in Table~\ref{sup:tab:sk_acc_miou_all_splits}(c).  
We achieve an HmIoU of only $1.4\%$. While the accuracy of \class{Motorcyclist} is quite high, the very low IoU shows that it is very hard to distinguish the different unseen classes and a lot of examples are wrongly classified as \class{Motorcyclist}.

\subsection{Visualisation of class prototype spaces}
\begin{figure*}[!htb]
\centering
\includegraphics[width=\linewidth]{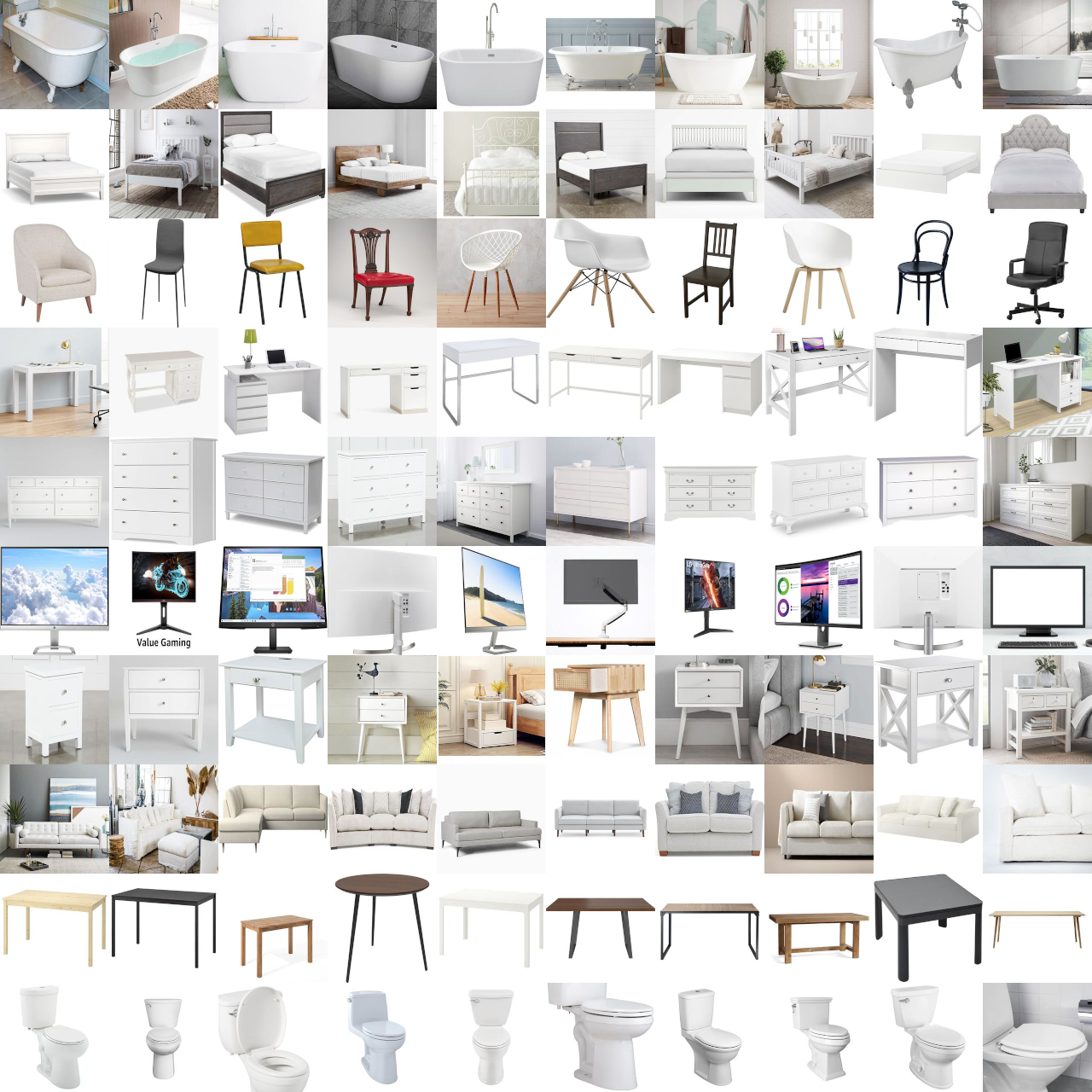}
\vspace*{-2mm}
\caption{The first ten images collected with Google Images and used to generate the image embeddings of ModelNet40 (unseen classes).}
\label{supp:fig:uneen_images_google}
\vspace{-1mm}
\end{figure*}

To assess the difficulty of transferring knowledge of seen classes to unseen classes, we examine the textual semantic similarities and dissimilarities of word embeddings. To that end, we provide in Figure~\ref{supp:fig:ss_tsne_representation} t-SNE visualizations \cite{van2014accelerating} of the W2V+GloVe class prototypes for the semantic segmentation datasets (S3DIS, ScanNet, SemanticKITTI). These diagrams can be compared to the respective performance Tables~\ref{sup:tab:s3dis_acc_miou}, \ref{sup:tab:sn_acc_miou}, \ref{sup:tab:sk_acc_miou_all_splits} and confusions matrices in Figures \ref{supp:fig:seg_cm_s3dis}-\ref{supp:fig:seg_cm_sk}.

\subsection{Upper bound for semantic segmentation}

\begin{table}[t]
\centering\setlength{\tabcolsep}{4pt}
\begin{tabular}{l|cc|c|cc}
\toprule
\multicolumn{1}{c|}{} &\multicolumn{2}{c|}{ZSL}& & \multicolumn{2}{c}{Supervised} \\ 
\cmidrule{2-6}
HM             & \stackbox{W2V\&\\Glove} & image & ``ideal'' & \stackbox{ZSL\\backb.} & \stackbox{Full\\superv.}\\
\toprule
S3DIS           & 12.9  & \hphantom{0}5.7   & 21.0 & 31.8 & 59.6 \\
ScanNet         & 12.5  & 15.5              & 17.0 & 40.3 & 47.2 \\
Sem.KITTI   & 17.1  & \hphantom{0}5.3   & 17.5 & 21.2 & 54.5 \\
\bottomrule
\end{tabular}
\vspace{2mm}
\caption{Comparing the ZSL results  with the upper bound (``ideal'') and with supervised models.}
\label{sup:tab:upper_bound}
\end{table}

Following the idea of a helpful anonymous reviewer, we experimented with 3D features as class prototypes, only to get a kind of upper bound as this does not satisfy the zero-shot principle. We trained the 3D backbone under full supervision using only seen classes. Then we created class prototypes using the ground-truth segmentation masks of seen and unseen classes, averaging features of all correctly-classified points for each seen class, and features of all points for each unseen class. Finally, we used our method with these ``ideal'' prototypes instead of the word- or image-based prototypes. Because these prototypes are obtained using knowledge about the 3D backbone, we had to reduce our bias term to zero. The results are summarized in the Table~\ref{sup:tab:upper_bound}. They suggest that we are probably close to the best results one can hope for with this kind of generative approach with that backbones.

\section{Image-based representations}
\label{sup:sec:image_representation}

To built our class prototypes based on image representations, we use images of objects belonging to each class of the datasets and extract deep features for each of these images using a pre-trained CNN on ImageNet \cite{russakovsky2015imagenet}. Details about the selection of the images are given in Section~\ref{sec:image_selection} and feature extraction is described in Section~\ref{sec:image_feature_extraction}. In Section~\ref{sec:image_classwise_results_ss} the classwise results for the different datasets are given and in Section~\ref{sec:sensitivity_image_quality} the sensitivity to the image collection quality is discussed.
\vspace{-1mm}

\subsection{Image selection}
\label{sec:image_selection}

\begin{table*}[!htbp]
\centering
\newcommand*\rotext{\multicolumn{1}{R{45}{1em}}}
\setlength{\tabcolsep}{3pt}
\scalebox{0.95}{\begin{tabular}{ll|rrrrrrrrr|>{\columncolor{LightGrey}}r>{\columncolor{LightGrey}}r>{\columncolor{LightGrey}}r>{\columncolor{LightGrey}}r|r}
\toprule
(a)&& \multicolumn{9}{c|}{seen classes} & \multicolumn{4}{>{\columncolor{LightGrey}}c|}{\textbf{unseen classes}} & \cellcolor{HL} \\
\multicolumn{1}{c}{\raisebox{6mm}{\stackbox{S3DIS}}} & \multicolumn{1}{c|}{\rotatebox{90}{super\rlap{vis.}}} & \rotext{Board} & \rotext{Bookcase} & \rotext{Ceiling} & \rotext{Chair} & \rotext{Clutter} & \rotext{Door} & \rotext{Floor} & \rotext{Table} & \rotext{Wall} & \rotext{\textbf{Beam}} & \rotext{\textbf{Column}} & \rotext{\textbf{Sofa}} & \rotext{\textbf{Window}} & \stackbox{Hm\\IoU} \\ \hline
W2V+Glove & self & 
19.1&   34.1&   92.8&   56.3&   39.2&
25.4&   91.5&   57.3&   62.3&   13.9&
2.4&    4.9& 8.1 & \cellcolor{HL} \bf 12.9\\ \hline
ResNet-18 \cite{he2016deep} & full & 
43.3&   36.5&   92.7&   67.7&   33.8&
45.9&   90.8&   62.9&   65.5&   0.8& 
0.7&    4.9&    5.5 & \cellcolor{HL} 5.7 \\
ResNet-50 \cite{caron2020unsupervised} & self  & 
25.0&   37.0&   93.5&   64.1&   36.8&
34.6&   91.4&   55.4&   65.4&   0.3&
1.0&    3.4&    3.9 & \cellcolor{HL} 4.1 \\
\bottomrule
\end{tabular}}

\vspace*{0.5mm}
\centering
\setlength{\tabcolsep}{2pt}
\scalebox{0.95}{\begin{tabular}{ll|rrrrrrrrrrrrrrrr|>{\columncolor{LightGrey}}r>{\columncolor{LightGrey}}r>{\columncolor{LightGrey}}r>{\columncolor{LightGrey}}r|r}
\toprule
(b)&& \multicolumn{16}{c|}{{seen classes}} &  \multicolumn{4}{>{\columncolor{LightGrey}}c|}{\textbf{unseen classes}} & \cellcolor{HL} \\
\multicolumn{1}{c}{\raisebox{9mm}{\stackbox{ScanNet}}} & \multicolumn{1}{c|}{\rotatebox{90}{super\rlap{vis.}}} & \rotext{Bathtub} & \rotext{Bed} & \rotext{Cabinet} & \rotext{Chair} & \rotext{Counter} & \rotext{Curtain} & \rotext{Door} & \rotext{Floor}  & \rotext{Other furniture} & \rotext{Picture} & \rotext{Refrigerator} & \rotext{Shower curtain} & \rotext{Sink} & \rotext{Table} & \rotext{Wall} & \rotext{Window} & \rotext{\textbf{Bookshelf}} & \rotext{\textbf{Desk}} & \rotext{\textbf{Sofa}} & \rotext{\textbf{Toilet}} & \stackbox{Hm\\IoU}\\
\hline
W2V+Glove & self & 
64.9&	44.0&	16.9&	63.2&	15.3&	33.8&	10.4&	91.0&	10.1&	4.3&	26.1&	0.2&	27.5&	43.1&	71.3&    2.8&	6.3&	3.3&	13.1&	8.1
& \cellcolor{HL} 12.5 \\ 
\hline
ResNet-18 \cite{he2016deep} & full & 
50.2&   40.6&  15.7&    61.1&   8.3&    32.9&
9.7&    90.9&   5.2&    0.8&    26.3&   0.1&
24.7&   43.8&   72.1&   3.8&    8.9&    15.6&
8.0&    3.7 & \cellcolor{HL} 13.9
\\
ResNet-50 \cite{caron2020unsupervised} & self & 
56.7&   42.7&   16.0&   59.4&   13.0&
35.0&   10.8&   90.9&   7.2&    0.4&
29.7&   11.9&   25.6&   40.2&   72.4&
3.4&    10.7&   15.7&   11.3&   3.1 & \cellcolor{HL} \bf 15.5
\\
\bottomrule
\end{tabular}}

\vspace*{0.5mm}
\centering
\setlength{\tabcolsep}{3pt}
\scalebox{0.95}{\begin{tabular}{ll|rrrrrrrrrrrrrrr|>{\columncolor{LightGrey}}r>{\columncolor{LightGrey}}r>{\columncolor{LightGrey}}r>{\columncolor{LightGrey}}r|r}
\toprule
(c)&& \multicolumn{15}{c|}{{seen classes}\raisebox{-2mm}{}} & \multicolumn{4}{>{\columncolor{LightGrey}}c|}{\textbf{unseen classes}} & \cellcolor{HL}  \\
\multicolumn{1}{c}{\raisebox{3.5mm}{\stackbox{Semantic\\KITTI\\ \footnotesize(main split, 1)}}}& \multicolumn{1}{c|}{\rotatebox{90}{super\rlap{vis.}}} & \rotext{Bicycle} & \rotext{Building} & \rotext{Car} & \rotext{Fence} & \rotext{Motorcyclist} & \rotext{Other ground} & \rotext{Other vehicle} & \rotext{Parking}  & \rotext{Person} & \rotext{Pole} & \rotext{Road} & \rotext{Sidewalk} & \rotext{Terrain} & \rotext{Trunk} & \rotext{Vegetation} & \rotext{\textbf{Bicyclist}} & \rotext{\textbf{Motorcycle}} & \rotext{\textbf{Traffic sign}} & \rotext{\textbf{Truck}} & \stackbox{Hm\\IoU}\\
\hline
W2V+Glove & self & 
0.0&    87.3&   86.9&   61.8&   0.0&
0.0&    0.0&    18.6&   0.0&    0.0&
88.8&   78.6&   73.6&   38.2&   87.8&
28.0&   11.5&   0.9&    2.6    & \cellcolor{HL} \bf 17.1 \\ 
\hline
ResNet-18 \cite{he2016deep} & full & 
0.0&    85.6&   93.3&   66.0&   0.0&
0.0&    0.0&    0.3&    0.0&    0.0&
87.7&   75.3&   70.3&   62.4&   87.4&
1.5&    0.3&    0.0&    5.7 & \cellcolor{HL} 3.6
\\
ResNet-50 \cite{caron2020unsupervised} & self & 
0.0&    86.4&   93.0&   61.4&   0.0&
0.0&    0.8&    7.2&    0.0&    0.0&
88.9&   77.9&   72.8&   62.3&   88.0&
4.1&    1.9&    0.0&    5.3  & \cellcolor{HL} 5.3 \\
\bottomrule
\end{tabular}}
\vspace*{0.5mm}
\caption{Classwise semantic segmentation performance (IoU in \%) on datasets S3DIS (a), ScanNet (b) and SemanticKITTI (c), with three different kinds of embeddings as class prototypes: (1)~W2V+GloVe word embeddings, (2)~image embeddings from a ResNet-18 fully supervised on ImageNet~\cite{he2016deep}, (3)~image embeddings from a ResNet-50 self-supervised on ImageNet~\cite{caron2020unsupervised}. Unseen classes are in bold face.}
\label{supp:tab:image_miou}
\vspace*{-2mm}
\end{table*}
For each of the classes, we collect the first 100 images returned by a Google image search with the corresponding class name, using the option to select images with a majority of white pixels. This search setting is used to favor the selection of images containing only one object, typically on a white background.
We show in Figure \ref{supp:fig:uneen_images_google} the first ten images obtained for the unseen classes of ModelNet40 using this procedure. It can be seen that most of these images indeed contain only one object of the desired class. 

Our reason for such a setting is that we would like the CNN to extract features that are specific to each object class, and also with less noise coming from background pixels. Note that this use of images in the wild comes with strictly no annotation effort, in the spirit of zero-shot learning. The paper actually reports results with a network pre-trained with self-supervision~\cite{He2020moco}, as well as with a network pre-trained with full supervision.

Please also note that, although ImageNet features a thousand classes, a number of seen and unseen classes in our datasets are not classified in ImageNet (which however does not mean they cannot appear in the background), e.g., seen \class{Ceiling} and \class{Floor}, and unseen \class{Column} in S3DIS; seen \class{Counter} and \class{Sink} in ScanNet; seen \class{Building} and \class{Road}, and unseen \class{Bicyclist} in SemanticKITTI.

In any case, even objects of classes that appear both among ImageNet categories and among the classes of our 3D datasets come with very different modalities, i.e., image vs point cloud. Besides, they are never directly associated as we only use the image-based pre-trained network to create embeddings from the images selected as described above.

\subsection{Construction of image-based representations}
\label{sec:image_feature_extraction}

For each of the 100 images that we collected for each class, as described in Section~\ref{sec:image_selection}, we extract the image features obtained after the global average pooling layer of the pre-trained CNN. These 100 image features are then averaged for each class before being $\ell_2$-normalised. This constructs the image-based class prototypes that we used for the ZSL and GZSL tasks (Section~4.5 and Table~3 of the paper).

\subsection{Classwise results for semantic segmentation}
\label{sec:image_classwise_results_ss}
\begin{figure*}[!t]
\centering
\vspace{15mm}
\begin{tabular}[t]{c|c}
ResNet-18 \cite{he2016deep} (full supervis.) & ResNet-50 \cite{caron2020unsupervised} (self-supervis.)
\\ \toprule
    \raisebox{-.5\height}{\includegraphics[width=0.47\linewidth,trim=110 30 0 20,clip]{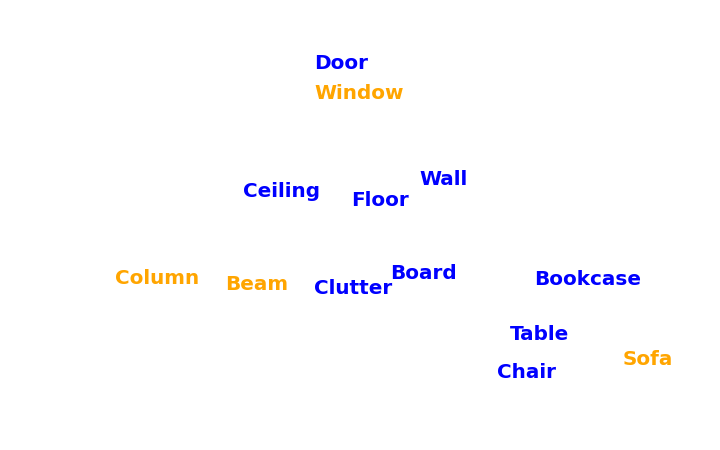}}&
    \raisebox{-.5\height}{\includegraphics[width=0.47\linewidth,trim=110 30 0 20,clip]{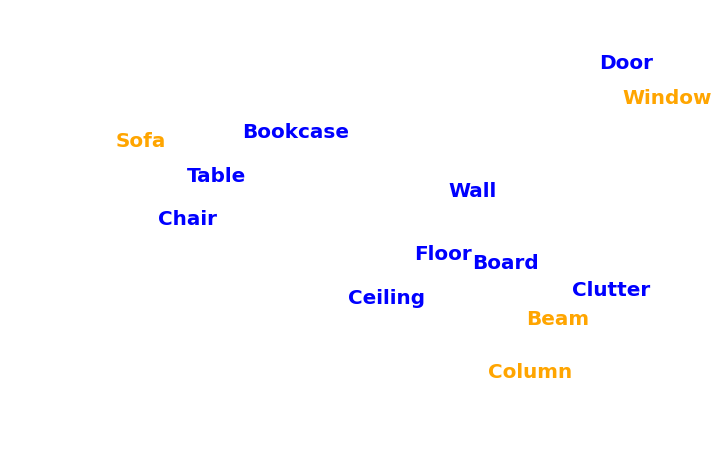}}
   \\ 
    \small (a) S3DIS  & \small (b) S3DIS
    \\ 
    \midrule
    \raisebox{-.5\height}{\includegraphics[width=0.47\linewidth,trim=110 30 0 20,clip]{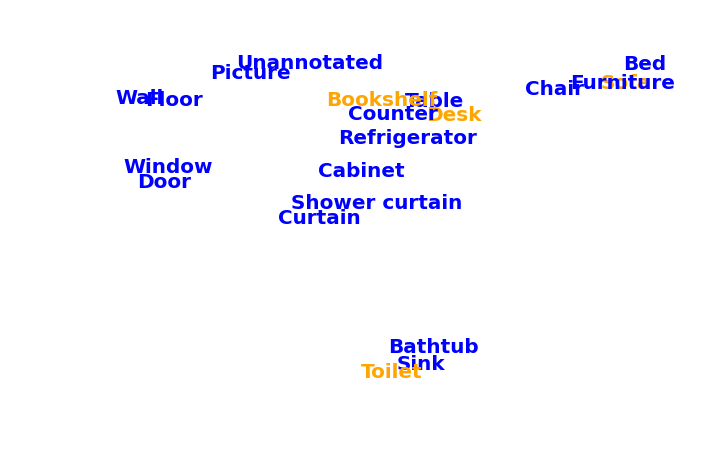}}&
    \raisebox{-.5\height}{\includegraphics[width=0.47\linewidth,trim=110 30 0 20,clip]{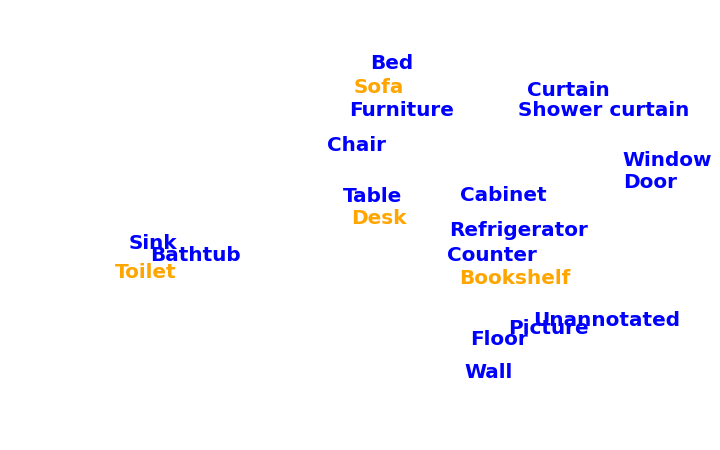}}
   \\
    \small (c) ScanNet  & \small (d) ScanNet  
      \\ \midrule
    \raisebox{-.5\height}{\includegraphics[width=0.47\linewidth,trim=110 30 0 20,clip]{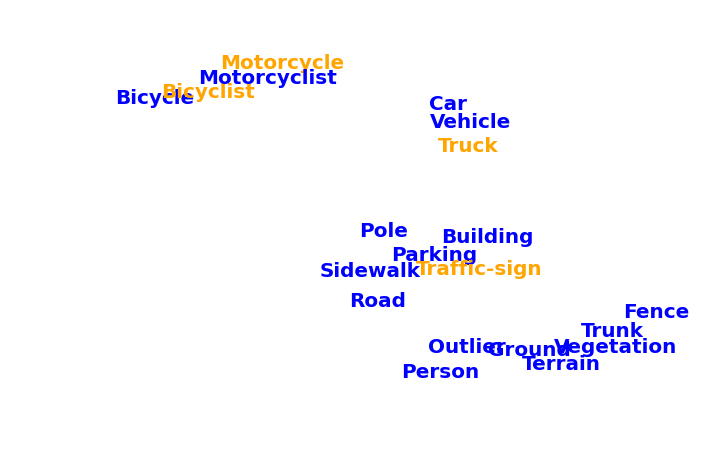}}&
    \raisebox{-.5\height}{\includegraphics[width=0.47\linewidth,trim=110 30 0 20,clip]{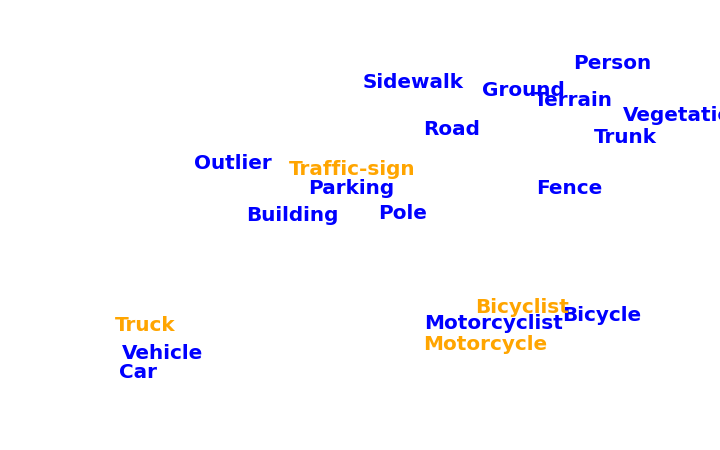}}
   \\
    \small (e) SemanticKITTI \ (main split) & \small (f) SemanticKITTI \ (main split)
    \\
   \bottomrule
\end{tabular}
\vspace*{2mm}
\caption{t-SNE \cite{van2014accelerating} visualizations of image-based embeddings for the semantic segmentation datasets (\textcolor{blue}{\bf blue seen}, \textcolor{orange}{\bf orange unseen}).}
\label{supp:fig:ss_image_tsne_representation}
\vspace{15mm}
\end{figure*}
We report in Table~\ref{supp:tab:image_miou} our classwise GZSL performance on datasets S3DIS, ScanNet and SemanticKITTI, when using either word embeddings or image embeddings. The word embeddings are the concatenation W2V+GloVe. The image-based representations are extracted using a ResNet-18 trained under full supervision on ImageNet \cite{he2016deep}, or a ResNet-50 trained by self-supervision on ImageNet \cite{caron2020unsupervised}.

With S3DIS, in Table \ref{supp:tab:image_miou}(a), we observe relatively similar results on seen classes, whether we use word or image embeddings. However, regarding unseen classes, we get a significantly lower IoU for classes \class{Beam} and \class{Column} using the image-based representation, compared to the word-based representation. We suppose that the quality of the retrieved images for these classes, mainly due to ambiguities, explains the poor performance. As a matter of fact, \class{Beam} images are disparate, containing, e.g., images of light beam or of the drink Jim Beam, and many of the \class{Column} images picture antique columns, or columns from Excel sheets..

Likewise, the results reported in Table~\ref{supp:tab:image_miou}(b) on ScanNet show that the performance is nearly independent of the type of class prototype on the seen classes. A possible explanation for the drop of performance on the class \class{Picture}, in particular with the image-based representations, could be a confusion between the scene that is pictured and the picture itself. We hypothesize that the good results for the unseen classes come from the large number of collected images which unambiguously display the unseen objects, as opposed to the S3DIS case.

Finally, on SemanticKITTI, we reach again similar performance for all types of class representations for most of the seen classes. Among the unseen classes, the IoU drops significantly for the classes \class{Bicyclist} and \class{Motorcycle} when using the image-based representations whereas it doubles for the class \class{Truck}. A possible explanation for the drop is that many images of the class \class{Motorcycle} actually shows someone riding the motorcycle, which is considered as class \class{Motorcyclist} in SemanticKITTI. A similar phenomenon is observed for classes \class{Bicycle} and \class{Bicyclist}. It also probably explain why the t-SNE representation of these pairs of classes are close to each other (see Figure~\ref{supp:fig:ss_image_tsne_representation}).

\subsection{Visualisation of class prototype spaces}

Figure~\ref{supp:fig:ss_image_tsne_representation} shows t-SNE visualizations \cite{van2014accelerating} of the class prototypes extracted for the three datasets.

We observe the same clusters for both kinds of pre-trained networks, which confirms that the difference in pre-training only has a somehow marginal impact on the results. (Please also remember that t-SNE visualization is not deterministic.)
Besides, these diagrams remain consistent with the groupings already observed with word embeddings (cf.\ Figure~\ref{supp:fig:ss_tsne_representation}), although they slightly differ.

\subsection{Sensitivity to the image collection quality}
\label{sec:sensitivity_image_quality}
To evaluate the impact of bad images in the image collections harvested automatically, we manually removed images that were not correct instances of the desired classes. As we are only removing images and not adding new ones, the image collections are smaller after this process.
Consequently, it may have both a positive and a negative effect.

Results are shown in Table~\ref{supp:tab:image_emb_opt}.
We notice a significant improvement on SemanticKITTI, where the HmIoU more than doubles. We also notice an improvement on the other datasets, except a slight drop of performance for ScanNet when using ResNet-50, possibly due to reduction of the number of images. This experiment confirms that finding images that unambiguously represents the object category is key in reaching a good performance.

\begin{table}[h]
    \centering
    \renewcommand{\arraystretch}{0.9}
    \setlength{\tabcolsep}{4pt}
    \scalebox{0.95}{\begin{tabular}{l|cc|cc}
    \toprule
    & \multicolumn{2}{c|}{\relax{ResNet-18 \cite{he2016deep}}} & \multicolumn{2}{c}{\relax{ResNet-50 \cite{caron2020unsupervised}}} \\
    & \multicolumn{2}{c|}{(full supervis.)} & \multicolumn{2}{c}{\relax{(self-supervis.)}} 
    \\
    Dataset  &  Original   & Denoised & Original & Denoised \\
    \midrule
    S3DIS  & ~~5.7  & \bf ~~6.5 & ~~4.1 & \bf ~~7.9 
    \\
    ScanNet & 13.9  & \bf 15.5 & \bf 15.5 & 14.7
    \\
    SemanticKITTI& ~~3.6  & \bf ~~8.2 & ~~5.3 & \bf 11.1
    \\
    \bottomrule
    \multicolumn{1}{l}{}\\[-7pt]
    \end{tabular}}
    \vspace*{1mm}
    \caption{Impact on HmIoU (\%) when ``denoising'' the image collections, using both kinds of pre-trained networks.}
    \label{supp:tab:image_emb_opt}
\end{table}

\fi

{\small
\bibliographystyle{ieee_fullname}
\bibliography{main}
}

\end{document}